\documentclass[lettersize,journal]{IEEEtran}
\usepackage{amsmath,amssymb,amsfonts}
\usepackage{algorithmic}
\usepackage{algorithm}
\usepackage{array}
\usepackage[caption=false,font=normalsize,labelfont=sf,textfont=sf]{subfig}
\usepackage{textcomp}
\usepackage{stfloats}
\usepackage{url}
\usepackage{verbatim}
\usepackage{graphicx}
\usepackage{cite}
\usepackage{multirow}
\usepackage{threeparttable}
\usepackage{hyperref}
\usepackage{siunitx}
\usepackage{xcolor}

\def\BibTeX{{\rm B\kern-.05em{\sc i\kern-.025em b}\kern-.08em
    T\kern-.1667em\lower.7ex\hbox{E}\kern-.125emX}}

\usepackage[acronym]{glossaries}
\newacronym{ar}{AR}{Augmented Reality}
\newacronym{vr}{VR}{Virtual Reality}
\newacronym{mr}{MR}{Mixed Reality}
\newacronym{mse}{MSE}{Mean Squared Error}
\newacronym{drl}{DRL}{Deep Reinforcement Learning}
\newacronym{kc-td3}{KC-TD3}{Knowledge-assisted Constrained Twin-Delayed Deep Deterministic}
\newacronym{hci}{HCI}{Human-Computer Interaction}
\newacronym{cvar}{CVaR}{Conditional Value-at-Risk}
\newacronym{mtp}{MTP}{Motion-To-Photon}
\newacronym{embb}{eMBB}{Enhanced Mobile Broadband}
\newacronym{urllc}{URLLC}{Ultra-Reliable Low-Latency Communication}
\newacronym{mmtc}{mMTC}{Massive Machine Type Communication}
\newacronym{e2e}{E2E}{End-to-End}
\newacronym{aoi}{AoI}{Age of Information}
\newacronym{iiot}{IIoT}{Industrial Internet of Things}
\newacronym{mi}{MI}{Mutual Information}
\newacronym{lstm}{LSTM}{Long Short-Term Memory}
\newacronym{mlp}{MLP}{Multi-Layer Perception}
\newacronym{hdmi}{HDMI}{High-Definition Multimedia Interface}
\newacronym{apdo}{APDO}{accelerated primal-dual policy optimization}
\newacronym{cmdp}{CMDP}{Constrained Markov Decision Processes}
\newacronym{lstf}{LSTF}{Long Sequence Time-series Forecast}
\newacronym{vpg}{VPG}{Vanilla Policy Gradient}
\newacronym{trpo}{TRPO}{Trust Region Policy Optimization}
\newacronym{ddpg}{DDPG}{Deep Deterministic Policy Gradient}
\newacronym{ppo}{PPO}{Proximal Policy Optimization}
\newacronym{rmp}{RMP}{Riemannian Motion Policy}
\newacronym{var}{VaR}{Value-at-Risk}
\newacronym{kpi}{KPI}{Key Performance Indicator}
\newacronym{dt}{DT}{Digital twin}
\newacronym{agv}{AGV}{Autonomous Guided Vehicle}
\newacronym{isac}{ISAC}{Integral sensing and communication}
\newacronym{pd}{PD}{proportional–derivative} 
\newacronym{gae}{GAE}{ Generalized Advantage Estimate}
\newacronym{ccdf}{CCDF}{Complementary Cumulative Distribution Function}
\newacronym{mam}{MA}{Moving Average Method}
\newacronym{dof}{DoF}{degrees of freedom}
\newacronym{ofdm}{OFDM}{Orthogonal Frequency-Division Multiplexing}
\newacronym{iot}{IoT}{Internet of Things}
\newacronym{dh}{D-H}{Denavit–Hartenberg}
\newacronym{it}{IT}{Information Technology}
\newacronym{dnn}{DNN}{Deep Neural Network}
\newacronym{arma}{ARMA}{Autoregressive Moving Average}
\newacronym{tcp}{TCP}{Transmission Control Protocol}
\newacronym{rpc}{RPC}{Remote Procedure Call}
\newacronym{tc}{TC}{Traffic Control}
\newacronym{rmse}{RMSE}{Root Mean Squared Error}
\newacronym{pid}{PID}{Proportional Integral Derivative}
\newacronym{aol}{AoL}{Age of Loop}
\newacronym{voi}{VoI}{value of information}
\newacronym{ros}{ROS}{Robot Operating System}
\newacronym{maml}{MAML}{Model-Agnostic Meta-Learning}
\newacronym{cps}{CPS}{Cyber-Physical System}
\newacronym{hitl}{HITL}{human-in-the-loop}
\newacronym{csv}{CSV}{Comma Separated Values}
\newacronym{ai}{AI}{Artificial Intelligence}
\newacronym{rl}{RL}{Reinforcement Learning}
\newacronym{qoe}{QoE}{quality of experience}
\newacronym{fov}{FoV}{field of view}
\newacronym{rnn}{RNN}{Recurrent Neural Network}
\newacronym{mec}{MEC}{mobile edge computing}
\newacronym{llm}{LLM}{large language model}
\newacronym{sew}{SEW}{Smart Energy Water}
\newacronym{ugv}{UGV}{Unmanned Ground Vehicle}
\newacronym{uav}{UAV}{Unmanned Aerial Vehicle}
\newacronym{wncs}{WNCS}{Wireless Networked Control System}
\newacronym{api}{API}{Application Programming Interface}
\newacronym{hm}{HITL-MAML}{Human-In-The-Loop Model-Agnostic Meta-Learning}
\newacronym{sgd}{SGD}{Stochastic Gradient Descent}
\newacronym{psnr}{PSNR}{Peak Signal-to-Noise Ratio}
\newacronym{ssim}{SSIM}{Structural Similarity Index}
\newacronym{lpips}{LPIPS}{Learned Perceptual Image Patch Similarity}
\newacronym{jet}{JET}{Joint European Torus}
\newacronym{sacgcn}{SAC-GCN}{Soft Actor–Critic and Graph Convolutional Networks}
\newacronym{vsp}{VSP}{Virtual Service Provider}
\newcommand{\traj}{\mathcal{T}}

\usepackage{cleveref}
\crefname{table}{Table}{Tables}
\crefname{figure}{Fig.}{Figs.}
\crefname{section}{Section}{Sections}

\begin{document}

\title{Task-Oriented Edge-Assisted Cross-System Design for Real-Time Human-Robot Interaction in Industrial Metaverse}

\author{Kan~Chen, \IEEEmembership{Student Member, IEEE}, Zhen~Meng, Xiangmin~Xu, \IEEEmembership{Student Member, IEEE}, Jiaming~Yang, \IEEEmembership{Student Member}, Emma Li, \IEEEmembership{Member, IEEE}, and Philip G. Zhao, \IEEEmembership{Senior Member, IEEE}
\thanks{K. Chen, Z. Meng, X. Xu, and E. Li are with the School of Computing Science, University of Glasgow, UK. 
(e-mail: \{k.chen.1, x.xu.1, j.yang.9\}@research.gla.ac.uk; \{zhen.meng, liying.li\}@glasgow.ac.uk)}
\thanks{P. G. Zhao is with the Department of Computer Science, University of Manchester, UK. (e-mail: philip.zhao@manchester.ac.uk)}
\thanks{Corresponding author: Zhen Meng.}
}

\markboth{IEEE Transactions on Mobile Computing, ~Vol.~/, No.~/, Augest~2025}%
{}

\maketitle

\begin{abstract}


Real-time human-device interaction in industrial Metaverse faces challenges such as high computational load, limited bandwidth, and strict latency. This paper proposes a task-oriented edge-assisted cross-system framework using digital twins (DTs) to enable responsive interactions. By predicting operator motions, the system supports: 1) proactive Metaverse rendering for visual feedback, and 2) preemptive control of remote devices. The DTs are decoupled into two virtual functions—visual display and robotic control—optimizing both performance and adaptability. To enhance generalizability, we introduce the Human-In-The-Loop Model-Agnostic Meta-Learning (HITL-MAML) algorithm, which dynamically adjusts prediction horizons. Evaluation on two tasks demonstrates the framework’s effectiveness: in a Trajectory-Based Drawing Control task, it reduces weighted RMSE from 0.0712 m to 0.0101 m; in a real-time 3D scene representation task for nuclear decommissioning, it achieves a PSNR of 22.11, SSIM of 0.8729, and LPIPS of 0.1298. These results show the framework’s capability to ensure spatial precision and visual fidelity in real-time, high-risk industrial environments.

\end{abstract}


\section{Introduction}
\label{sec:introduction}
Industrial Metaverse represents an integrated virtual ecosystem that extends the concept of the Metaverse to specific industrial sectors, merging physical and digital realms. It explores the transformative potential of teleoperation, real-time collaboration, and synchronization within high-risk industries, driving substantial advancements in industrial operations~\cite{cao2023toward}. \glspl{dt} are a key enabler within the larger framework of industrial Metaverse, facilitating real-time data interaction and providing highly accurate virtual models of physical assets~\cite{glaessgen2012digital}. For instance, in nuclear decommissioning, industrial Metaverse, aided by \glspl{dt}, can facilitate real-time remote teleoperation to inspect critical components like internal wall tiles in fusion reactors. This enables precise monitoring and safe handling during dismantling, ensuring both safety and efficiency~\cite{pacheco2021multiple}~\cite{nuclear}.

\begin{figure}
\centering
\includegraphics[scale=0.32]{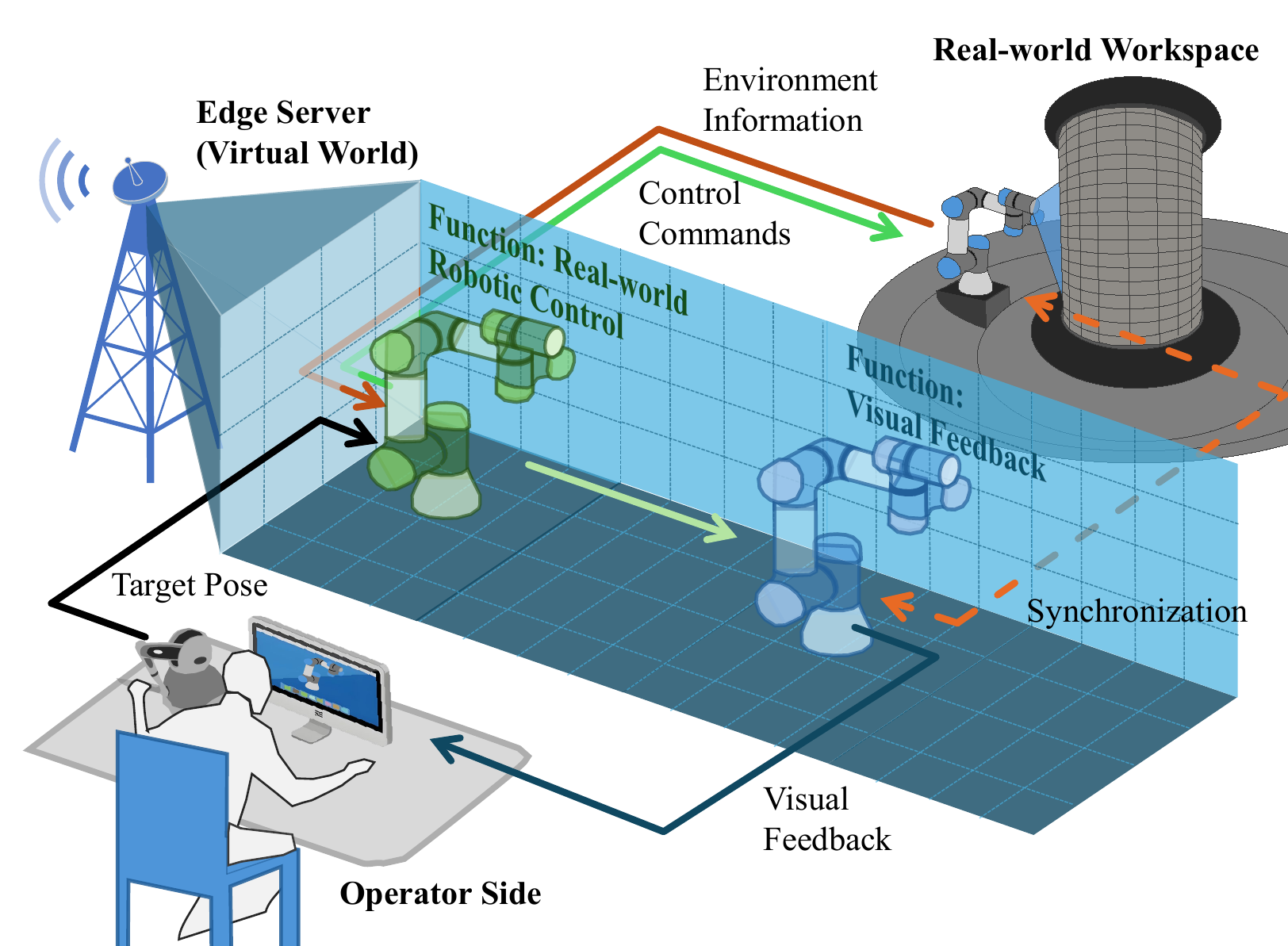}
\caption{Proposed task-oriented cross-system design framework.}
\label{Illustration of system model}
\end{figure}

Despite the advanced capabilities of industrial Metaverse, significant challenges remain in realizing its full potential. One of the fundamental issues lies in the hidden and dynamic nature of latency within highly coupled cyber-physical subsystems~\cite{9834918}. Unlike conventional networks where latency can be measured and managed in relative isolation, in industrial Metaverse environments latency emerges across multiple tightly interconnected subsystems (e.g., sensing, control, rendering, and actuation). This interdependence makes latency not only more difficult to analyze but also highly dynamic, since variations in one subsystem may propagate and amplify through others. Such complexity implicitly calls for cross-system design principles rather than subsystem-level optimization alone~\cite{10422886}.  

Compounding this challenge, the requirements of latency and reliability in industrial Metaverse are still poorly understood. Existing \glspl{kpi}, such as those established for ultra-low latency, high reliability, and high bandwidth, provide necessary technical baselines; however, they do not directly capture the task-level performance demands of industrial Metaverse applications. For instance, \cite{3GPP} defines fine-grained \glspl{kpi} for video streaming in industrial Metaverse scenarios, suggesting that data rates of 1 Gbps (for smooth playback) or 2.35 Gbps (for interactive experiences) are required to support 8K resolution, 120 FPS, and 360-degree video streaming~\cite{mangiante2017vr}. Similarly, the Tactile Internet prioritizes ultra-low latency to ensure that tactile sensations and control commands are transmitted almost instantaneously, with latency requirements often below 1 ms~\cite{7470948}, as humans are more sensitive to delays in tactile feedback than to audio or visual delays. This requirement is critical in applications where even slight feedback delays may cause errors or degraded performance, particularly in scenarios involving human-machine interaction or fine motor control~\cite{9203885}.  

Nevertheless, these technical metrics remain an imperfect proxy for actual task demands. Tasks in industrial Metaverse are often diverse and complex, ranging from collaborative teleoperation to predictive maintenance, each with distinct sensitivity to communication impairments~\cite{10105150}. Simply optimizing communication against conventional KPIs may therefore be suboptimal, as these metrics act like the ``shortest stave in the barrel'': the system must meet the most stringent requirement even if it is irrelevant to the task at hand. This mismatch reveals a critical gap between traditional communication metrics and task-specific requirements, motivating the development of task-oriented communication frameworks where communication resources are allocated and optimized directly toward task success~\cite{gunduz2022beyond}.  Furthermore, to reduce dependence on communication bottlenecks, edge deployment of virtual world components and decision-making modules is emerging as a promising direction~\cite{xu2022full}. By enabling intelligent decision-making closer to the physical processes, edge computing can reduce end-to-end latency and enhance real-time interaction between human operators, robots, and digital twins~\cite{10250875}. However, what functions should be deployed at the edge, how they should be partitioned between the edge and cloud, and how intelligent coordination can be achieved across heterogeneous nodes remain open and unresolved questions. Addressing these challenges requires not only innovations in networking protocols but also integrated approaches spanning communication, computation, and task-level system design.  

Furthermore, another significant challenge in industrial Metaverse is the intricate complexity of \gls{hitl} systems, where seamless interaction between human operators and automated systems is critical yet profoundly challenging. In these scenarios, delays in control commands and feedback loops can significantly affect system performance and safety~\cite{10466554}. For instance, in smart grids, where human operators manage and control critical infrastructure, latency in decision-making can lead to suboptimal responses to dynamic conditions such as load changes or fault detection~\cite{hitl}. While research on the effects of latency in teleoperation provides a foundational understanding~\cite{8283715}, the intricacies of human behavior and interaction with complex industrial Metaverse environments require more sophisticated models and simulations to predict and mitigate these effects. This presents a significant challenge for deploying \gls{ai}-driven control systems that can adapt to human behavior and ensure seamless, real-time operation.

\subsection{Related Work}
1) \textit{Edge-Assisted System for Industrial Metaverse:} By balancing the different communication, computation, and control capabilities of different devices and offloading some or all of the tasks to the edge or the cloud,
collaborative computing and sharing of multidimensional communication, caching, and computational resources including the \gls{llm} can be efficiently achieved in industrial \gls{cps}~\cite{10007839, 10521652, 10114989, 10639522, 9978919, 10648594, 10679152, hashash2023towards}. To tackle the challenge of coordinating the complexity and diversity of tasks, the authors in~\cite{10007839} proposed a task offloading scheme for \gls{mec} systems, minimizing deadline violation by using task migration, merging, and deep learning to optimize task execution order and reliability. 
To tackle the difficulty of agile troubleshooting and deployment of multi-source domain faults in industrial metaverse, the authors in~\cite{10521652} introduced a low-code intelligent fault diagnosis platform, integrating cloud-based infrastructure and edge-adaptive systems to enhance agile fault diagnosis, demonstrated on a \gls{sew} reduction platform and wind turbine generator. Furthermore, by exploiting the accurate digital representation of heterogeneous end-device and network state parameters in dynamic and complex \gls{iiot} scenarios in \gls{cps}, \gls{dt} models can be constructed at the edge to capture time-varying demands with low latency~\cite{10114989}. The authors in~\cite{10639522} proposed a cloud-edge-end collaborative intelligent service computation offloading scheme based on the \gls{dt} driven Edge Coalition Formation methodology to improve the offloading efficiency and utility of edge servers. efficiency and total utility.  On the other hand, to ensure accurate and scalable synchronization of industrial Metaverse, \cite{hashash2023towards} proposed a decentralized architecture that partitions the Metaverse into sub-metaverses hosted at the edge, each synchronized with corresponding digital twins using an optimal transport-based assignment scheme. 

2) \textit{Cross-System Design for Complex Systems:} Significant contributions have been made in the area of cross-system system design by jointly considering the communication, computing, and Control~\cite{9729746, 9305697, 8902186, 9268977}. The authors in~\cite{9729746} introduced an integrated scheduling method for sensing, communication, and control in \gls{uav} networks using mmWave/THz communications, enhancing backhaul data transmission. They analyzed the interaction between sensing and motion control, introducing a ``state-to-noise-ratio" concept linking control patterns to data rates. The authors in~\cite{9305697} optimized the data transmission in \gls{iiot} systems, which derived the average \gls{aoi} expression under packet loss and finite retransmissions, linking it to control performance and communication energy consumption. The authors in~\cite{8902186} considered \gls{urllc} scenarios, where mobile devices predict future states and send them to a data center. They optimized resources to improve delay-reliability tradeoffs, accounting for prediction errors and packet losses.
The authors in~\cite{9268977} introduced a proactive tile-based video streaming method for wireless \gls{vr} to reduce \gls{mtp} latency by predicting and delivering tiles before playback. However, most of these efforts are not task-oriented, resulting in a gap between traditional metrics for communications and the \glspl{kpi}. In addition, these efforts do not directly consider the interaction between the human and the environment in industrial Metaverse. 

3) \textit{Task-Oriented Communications Design for Intelligent Device:}
Unlike traditional methods that measure success based on bit error rates, task-oriented communication focuses on achieving the final task as the measure of success, which has been growing applied to a wide range of industrial Metaverse including \gls{iiot}~\cite{10437251}, ~\gls{ugv} and ~\gls{uav} communication systems~\cite{10634104} and industrial Metaverse~\cite{Jsac_zhen}. In~\cite{10437251}, the authors considered a \gls{wncs} scenario, where a goal-oriented scheduling and control co-design policy is proposed to minimize violation probability by prioritizing goal-relevant information under resource constraints. In~\cite{10634104}, the authors considered \gls{uav} tracking scenarios, where a goal-oriented communication framework using \gls{drl} is proposed to optimize control and communication data selection and repetition.
The authors in~\cite{Jsac_zhen} developed a task-oriented cross-system design framework to minimize packet rates for accurate robotic arm modeling in the Metaverse. However, the robotic arms in the Metaverse simply act as ``Digital Shadows" and do not have the ability to synthesize information from the environment as well as user inputs to make decisions on their own accord~\cite{sepasgozar2021differentiating}. In addition, the system design was based on offline data training and testing and did not directly address the latency problem of real-time systems with a human in the loop. The author in~\cite{chen2024real} developed a similar framework for the Metaverse, but it was only optimized based on a single task, and did not consider generalizability and robustness across multiple tasks.


\subsection{Contributions}

In this paper, we aim to solve the following problems: 1) How to eliminate the effects of time delay among different subsystem components to achieve real-time interactions? 2) How to design a strategy to leverage user behavior to address real-time \gls{hitl} interactions, while ensuring the algorithm's generalizability and robustness across various robotic arm operating behaviors? 3) How to build the prototype to verify the effectiveness of the proposed algorithm to ensure a reliable and adaptive industrial Metaverse operation? 
The main contributions of this paper are summarized as follows:

\begin{itemize} \item We propose a task-oriented cross-system design framework for real-time human–robot interaction in industrial Metaverse, tailored for remote inspection and manipulation in nuclear decommissioning. The framework jointly optimizes sensing, communication, prediction, control, and rendering by decoupling the digital twin into proactive visual feedback and predictive control modules.

\item We develop a behavior-aware horizon management algorithm that dynamically adjusts prediction windows for rendering and control. A two-stage learning process combines offline \gls{maml}-based pretraining and online \gls{hitl} adaptation, enhancing robustness in precision tasks such as tile inspection and component dismantling.

\item We implement a complete industrial Metaverse prototype integrating haptic input, visual feedback, a virtual twin, and a physical robotic arm for a simulated reactor tile inspection task. Experiments under varying communication delays show that our method improves accuracy and responsiveness over baseline approaches, demonstrating its effectiveness for real-world nuclear maintenance operations. \end{itemize}

The rest of this paper is organized as follows. In \cref{sec:method}, we propose the task-oriented cross-system design framework where all subsystems, i.e., sensing, communication, prediction, control, and rendering, are elaborated in detail.
In \cref{sec:hitl-mr}, we first briefly introduce the background of \gls{maml} and introduce our two-stage training algorithm.
\cref{sec:prototype} consists of a description of our prototype, as well as the data collection process. 
\cref{sec:results} provides the training settings and performance evaluations.
Finally, \cref{sec:conclusions} concludes this paper.

\section{Task-Oriented Cross-System Design}\label{sec:method}

In this section, we propose a task-oriented cross-system design framework. We introduce two functions of the framework, different sub-modules, the time flow chart, and the time delay composition.

\subsection{System Overview}


The proposed framework, illustrated in Fig.~\ref{Illustration of system model}, is designed to enable real-time human–robot interaction for remote inspection and manipulation tasks in nuclear decommissioning environments. It comprises three spatially distributed components: 1) an operator interface equipped with a haptic input device and a visual feedback display; 2) a virtual environment hosted on an edge server, containing a \gls{dt} of the robotic arm and a simulated workspace; and 3) a physical environment located within a nuclear fusion reactor vessel, where a real robotic arm operates to carry out inspection and dismantling tasks. These components communicate via a wireless network, simulating the constrained and high-latency communication conditions typically encountered in nuclear fusion facilities. During operation, the human operator observes the \gls{dt} and its surrounding environment via the visual interface and issues control commands using the input device. The \gls{dt} module integrates the operator’s intent, virtual environment feedback (e.g., simulated collisions or geometry), and real-time data from the physical robot to generate predictive control instructions. These commands are transmitted to the robotic arm deployed inside the reactor vessel, where it performs precision tasks such as in-vessel tile inspection, alignment, or removal. The resulting physical interactions are sensed, transmitted back to the \gls{dt}, and displayed to the operator, forming a closed-loop control system.


However, due to time delays introduced by the subsystems including sampling, computation, control, communication and rendering, discrepancies arise between the perceived trajectories of the \gls{dt} of the robotic arm, the real-world robotic arm, and the operator's input commands. Even worse, in this \gls{hitl} industrial Metaverse, time delays in visual feedback can lead to delays in the operator's control, and delays in control can further affect the generation of visual feedback, which finally significantly affects the robotic arm's performance of the task. To tackle this problem, the system predicts the operator's commands and dynamically changes the prediction horizons in real time. This prediction serves two main purposes: (1) to compensate for the delay between the operator’s motion and the visual feedback rendering of the \gls{dt}, and (2) to minimize the latency between the operator’s control inputs and the actions of the real-world robotic arm in the real-world environment.

\subsection{Information Flow}
\begin{figure}
\centering
\includegraphics[scale=0.40]{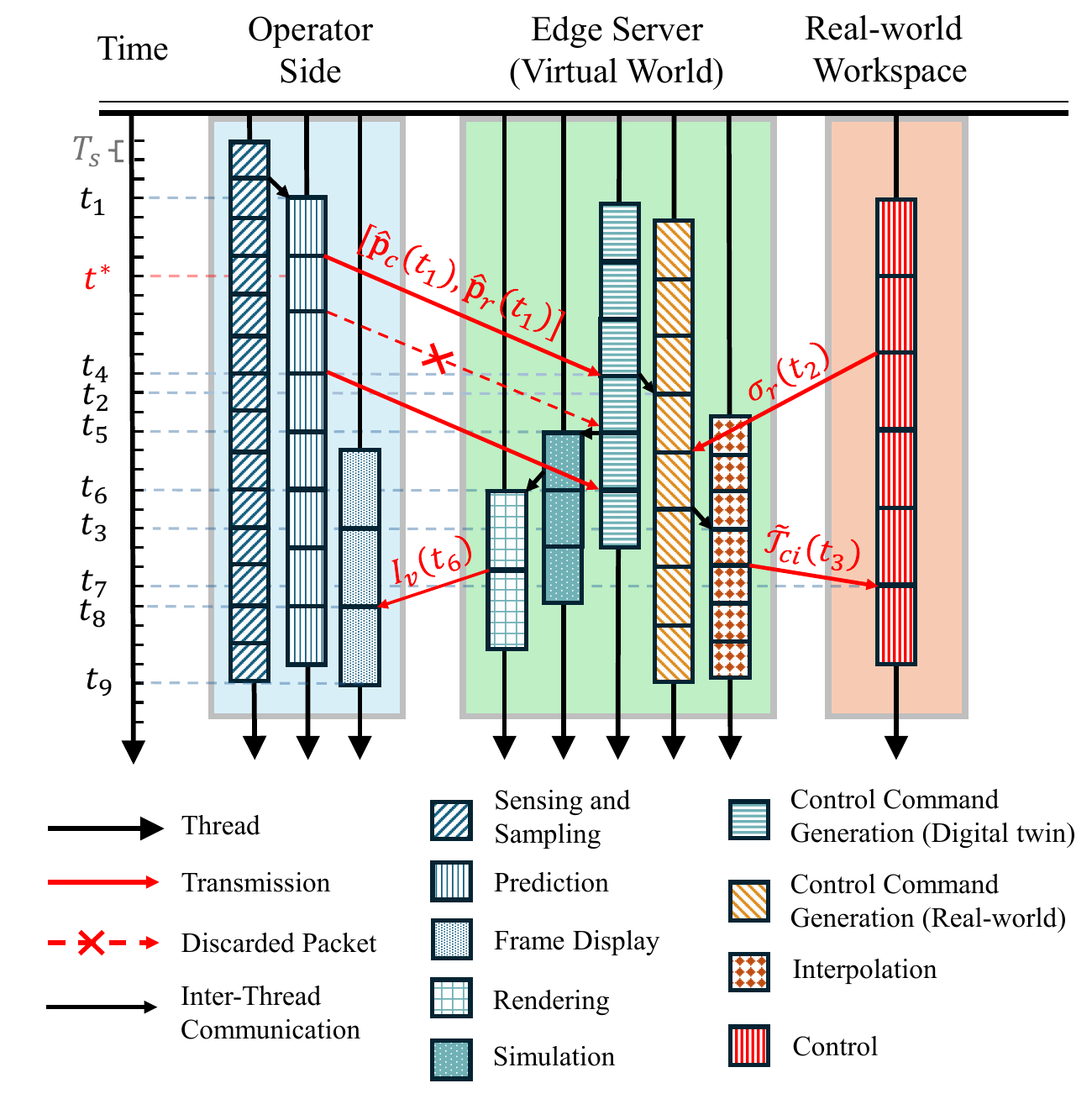}
\caption{Illustration of the information flow.}
\label{fig: time flow}
\end{figure}

As shown in Fig.~\ref{fig: time flow}, time is divided into time slots. We modified the entire working process queue as $M/M/1/2^*$~\cite{7415972}. In this case, only one packet is kept in the buffer waiting for processing, and packets waiting for processing are replaced upon the arrival of a more up-to-date packet. No packets are blocked from entering the queue. If a new packet arrives while the system is full, the packet waiting is discarded. The proposed framework comprises two information flows served for two functions: 1) real-world robotic control and 2) visual feedback.

The function of real-world robotic control, encompasses all processes from sensing the movement of the operator through data transmission, the Metaverse data processing, and real-world robotic control. During the process, the operator operates the input device and generates a series of poses $\{{\bf{p}}_\text{1}, {\bf{p}}_2,...,{\bf{p}}_i\}$ which are sampled by the built-in sensor. Next, In the $t_1$-th time slot, an agent decides the length of the prediction horizon. followed by a predictor generates the predicted pose for the visual feedback $\hat{{\bf{p}}}_v(t_1)$ and real-world robotic control $\hat{\bf{p}}_r(t_1)$. Then, $\hat{{\bf{p}}}_v(t_1)$ and $\hat{{\bf{p}}}_r(t_1)$ are transmitted to industrial Metaverse via networks. After that, in the $t_2$-th time slot, with the feedback of the environment, the received predicted poses $\hat{{\bf{p}}}_r(t_2)$ is processed to generate a control command in industrial Metaverse, and the Metaverse updates based on the corresponding command and the feedback of the environment. In the $t_3$-th time slot, to maintain smoothing motion trajectories in real-world workspace, the command is further interpolated to generate a joint position control command and transmitted via a communication channel for actuation. Finally, in the $t_8$-th time slot, the command is received by the real-world robotic arm and further transformed into low-level commands that can be directly executed. Considering the completed real-world robotic control workflow, we define the \gls{e2e} delay for the real-world robotic control by:
\begin{align}\label{eq: control latency}
   T_{r} =  t_{9} - t_1.
\end{align}
On the other hand, the function of the visual feedback, encompasses all processes from the sensing of movement of the operator, data transmission, data processing, rendering, and feedback display. Specifically, we repeat the steps before the process of those in the robotic control function. In addition, in the $t_4$-th time slot, the received predicted poses $\hat{{\bf{p}}}_v(t_4)$ is processed to generate a control command, After that, in the $t_5$-th time slot, the update in the Metaverse is simulated based on the corresponding command and the feedback of the environment. Next, in the $t_6$-th time slot, the virtual environment is rendered. Finally, in the $t_7$-th time slot, the rendered image ${\textbf{I}}_v(t)$ is streamed back to the user side. Similarly, we define the \gls{e2e} delay for the visual feedback by:
\begin{align}\label{eq: rendering latency}
   T_{v} =  t_{10} - t_1.
\end{align}

\subsection{System Components}

\subsubsection{Operator Side}
On the operator side, based on the position of the robot arm in the visual display, the operator controls the input devices to issue control commands
$\mathbf{p}(t)=[l_\mathrm{x}(t), l_\mathrm{y}(t), l_\mathrm{z}(t), q_\mathrm{x}(t), q_\mathrm{y}(t), q_\mathrm{z}(t), q_\mathrm{w}(t)]$, where $[l_\mathrm{x}(t), l_\mathrm{y}(t), l_\mathrm{z}(t)]$ is the desired position of in the Cartesian coordinate system and $[q_\mathrm{x}(t), q_\mathrm{y}(t), q_\mathrm{z}(t), q_\mathrm{w}(t)]$ is the desired orientation in quaternion~\cite{kuipers1999quaternions}. To compensate the time delay latent in all the subsystems, we introduce the predictor to predict the operator's pose which is used to 1) proactive rendering of virtual environments and 2) preemptive generation of control commands for the real robotic arm. Specifically, in the $t_1$-th time slot, given historical poses $\mathcal{P}(t_1) = [{\bf{p}}(t_1-W_p), {\bf{p}}(t_1-W_p+1), ...,{\bf{p}}(t_1)]$, the predictor predicts the pose for the real-world robotic arm control $\hat{\bf{p}}_{\it{r}}(t_1)$, and the proactive rendering $\hat{\bf{p}}_{\it{v}}(t_1)$, respectively. This is expressed by
\begin{align}\label{eq: prediction}
[\hat{\bf{p}}_{\it{r}}(t_1),&\hat{\bf{p}}_{\it{v}}(t_1)] = \mathcal{F}_p(\mathcal{P}(t_1), \theta_{p},H_r(t_1), H_v(t_1)),
\end{align}
where $\theta_{p}$ denotes the parameters of the prediction. $H_c(t_1)$ and $H_r(t_1)$ represent the lengths of prediction horizons. 
To illustrate, we propose to use~\gls{arma} for its low computational complexity, robustness, and no need for pre-training~\cite{choi2012arma}, but the framework works equally well with other predictors. Specifically, in the $t_1$-th time slot, The one-step predicted value \( \hat{\mathbf{p}}(t_1) \) is expressed as a combination of Auto-Regressive and Moving-Average components, formulated as follows:
\begin{align}
\hat{\mathbf{p}}(t_1) = c + \sum_{a=1}^p \phi_a \mathbf{p}(t_1 - a) + \sum_{b=1}^q \theta_b \epsilon(t_1 - b),
\end{align}
where $\hat{\mathbf{p}}(t_1)$ is the predicted value at time $t_1$, $c$ is the constant term that allows the model to fit time series data with a non-zero mean, $\phi_a$ and $\theta_b$ are the coefficients of the Auto-Regressive and Moving-Average components, respectively, and $\epsilon(t_1 - b)$ denotes the prediction errors. The Auto-Regressive term $\sum_{a=1}^p \phi_a \mathbf{p}(t_1 - a)$ captures the influence of the $p$-lagged past values of $\mathbf{p}(t_1)$, while the Moving-Average term $\sum_{b=1}^q \theta_b \epsilon(t_1 - b)$ models the effects of past forecast errors up to a lag of $q$. By recursively applying the \gls{arma} model over $H$ steps, we create a sequence of future predictions based on the current and previous observations.

In addition, to dynamically capture the time delay characters in different subsystems, we introduce a \gls{drl} agent that adaptively decides the prediction length $H_r(t_1)$ and $H_v(t_1)$, which is expressed by
\begin{align}
[H_r(t_1), H_v(t_1)] = \pi_\theta([{\bf{p}}(t_1), T_r(t_1), T_v(t_1)]),
\end{align}
where $T_r(t_1), T_v(t_1)$ are the measured \gls{e2e} delay defined by $T_{r} =  t_{9} - t_1$ and $T_{v} =  t_{10} - t_1$.

\subsubsection{The Metaverse}

The virtual environment, deployed on an edge server, generates 1)control commands for the real-world robotic arm and 2)visual feedback for the operator. At time slot $t_2$, the \gls{dt} synthesizes operator-predicted commands $[\hat{{\bf{p}}}_r(t_1), \hat{{\bf{p}}}_v(t_1)]$, the last simulation state $\sigma_v(t_2)$, and real-world feedback $\sigma_r(t_2)$ to compute control instructions:  
\begin{align}
\widetilde{\traj}_r(t_2) = \mathcal{F}_{r}(\hat{{\bf{p}}}_c(t_2), \sigma_v(t_2), \sigma_r(t_2),\theta_{c}),
\end{align}  
where $\theta_c$ represents control parameters, and $\widetilde{\tau}_i(t_2)$ is the $i$-th joint angle at $t_2$.  

To ensure smooth motion, we employ Riemannian Motion Policies (RMPFlow), a method merging local strategies into a global policy~\cite{ratliff2018riemannian}. The desired acceleration for each joint is:  
\begin{align}
\ddot{{\bf{q}}}_i(t_2) = k_p \cdot \mathcal{R}({\bf{q}}'_i(t_2) - {\bf{q}}_i(t_2)) - k_d \cdot \dot{{\bf{q}}}_i(t_2),
\end{align}  
where ${\bf{q}}_i(t_2)$ is the current joint angle, ${\bf{q}}'_i(t_2)$ the target position, $k_p$ the position gain, and $k_d$ the damping gain. The robust capping function $\mathcal{R}(\cdot)$ prevents excessive corrections:  
\begin{align}
\mathcal{R}({\bf{u}}(t_2)) =  
\begin{cases}  
{\bf{u}}(t_2), & \text{if } ||{\bf{u}}(t_2)|| < \theta_{th}, \\  
\theta_{th} \cdot \frac{{\bf{u}}(t_2)}{||{\bf{u}}(t_2)||}, & \text{otherwise}.
\end{cases}
\end{align}  

Since real-world execution requires control adaptation, we apply trajectory smoothing via linear interpolation~\cite{sim_rmp}:  
\begin{align}
{\bf{\hat{q}}}_i(t_3) =  
\begin{cases}  
\alpha_s \cdot {\bf{\hat{q}}}_i(t_3 - T_{co}) + (1 - \alpha_s) \cdot {\bf{q}}_i(t_3), & t_e < t_d \\  
{\bf{\hat{q}}}_i(t_3 - T_{co}), & \text{otherwise}.
\end{cases}  
\end{align}  
Here, $T_{co}$ is the interval between interpolated commands, and $\alpha_s$ evolves as $\alpha_s' = \alpha_s^2$.  

For visual feedback, the \gls{dt} updates its virtual model at $t_4$ using operator commands $\hat{{\bf{p}}}_r(t_4)$, the simulation state $\sigma_v(t_4)$, and real-world feedback $\sigma_r(t_4)$:  
\begin{align}
\widetilde{\traj}_v(t_4) = \mathcal{F}_{v}(\hat{{\bf{p}}}_r(t_4),\sigma_v(t_4), \sigma_r(t_4),\theta_{v}),
\end{align}  
where $\theta_v$ governs visual feedback functions. Simulation updates, including object interactions and collision detection, occur every $\Delta_t$ time steps:  
\begin{align}
\widetilde{\traj}_s(t_5) = \mathcal{F}_{u}(\widetilde{\traj}_v(t_5), \theta_{u}, \theta_{e}),
\end{align}  
where $\theta_u$ defines physics-based updates, and $\theta_e$ models robot-environment interactions.  

At $t_6$, the edge server renders the updated simulation:  
\begin{align}
{\bf{I}}_v(t_6) = \mathcal{F}_{re}(\widetilde{\traj}_s(t_6), \theta_{re}),
\end{align}  
where $\theta_{re}$ determines rendering parameters. The frame rate is given by $f_r = \frac{1}{T_{im}}$, where $T_{im}$ is the rendering time per frame. The final image ${\bf{I}}_v(t_6)$ is transmitted to the operator and displayed at $t_{10}$.

\subsubsection{Real-World Workspace}
In real-world workspace, the received motion control commands are further transformed into low-level commands that can be directly executed by the hardware by \glspl{api}. In the $t_7$-th time slot, the joint trajectory of the real-world robotic arm to be executed is expressed by
\begin{align}\label{eq: control_real}
\widetilde{\traj}_{rw}(t_7) = \mathcal{F}_{rw}(\widetilde{\traj}_{ri}(t_7),\theta_{rw}),
\end{align}
where $\mathcal{F}_{rw}(\cdot, \theta_{rw})$ represents the \gls{api}, and $\theta_{rw}$ denotes the corresponding parameters. 

To illustrate, we propose to use the~\gls{pid} control algorithm for its common ways to implement the control \gls{api}. Specifically, for the $i$-th joint of a robot, the joint position command in the $t_7$-th time slot is calculated by
\begin{align}
u_{i}(t_7) = K_{p}\tau_{i}(t_7) + K_{i}\int_{0}^{t_7}\tau_{i}(\dot{t}){\bf{d}}\dot{t} + K_{d}\frac{{\bf{d}}}{{\bf{d}}t}\tau_{i}(t_7),
\end{align}
where $K_{p}, K_{i}$ and $K_{d}$, all non-negative, denote the coefficients for the proportional, integral, and derivative terms, respectively, $\dot{t}$ represents the differentiation of $t$. The control is finished in the $t_{9}$-th time slot. Various sensors are also deployed in the real workspace to monitor the completion of the real-world robotic arm's tasks and feed the results back to the Metaverse on the edge server.

\subsection{Task Success Measurements}
To evaluate the effectiveness of the design, we conduct a rigorous assessment based on two dimensions: \textit{Trajectory Fidelity} and \textit{Task Accomplishment}. To this end, we have designed two types of remote interaction tasks:
\begin{itemize}
    \item \textit{Trajectory Tracking Task}: In this task, the operator is required to follow a predefined trajectory. The task performance is directly linked to the accuracy of the trajectory completion—the better the trajectory is followed, the higher the task performance. This task aims to assess the system’s real-time control capabilities and path accuracy, simulating remote operations such as surface following and contour scanning.
    \item \textit{Open-Ended Task}: Unlike the trajectory-based task, this task does not rely on specific paths. The operator is given a fixed amount of time to perform any operation, with the task’s success determined by the quality of the final result. This task evaluates the system’s responsiveness and its ability to handle more flexible, unstructured tasks within a time constraint, reflecting real-world scenarios where operations may vary in complexity.
\end{itemize}

These two tasks allow for a comprehensive evaluation of the system, from both the precision in task execution (trajectory tracking) and the quality of task outcomes (open-ended operation), ultimately validating the effectiveness of the remote interaction design under different operational conditions. Here is the detail:


\subsubsection{Trajectory Tracking Tasks-Drawing Shapes} 

In this task, the human operator manipulates the robotic arm via a haptic input device to trace predefined geometric patterns such as circles, triangles, stars, and squares. These trajectories emulate typical operations in remote handling, including precision surface following and contour scanning. The primary objective is to minimize the positional and orientational discrepancies between the intended trajectory and the executed path of the robotic arm, thereby reflecting the system's real-time responsiveness and motion fidelity.  To achieve accurate alignment between the operator's input and the robot’s execution, the input data must be transformed between different coordinate spaces. Since the input device and robotic arm operate in different workspaces, input values are mapped to the arm’s coordinate system. The end effector's pose, represented in Cartesian coordinates and quaternions, is transformed using a scaling matrix \( s \), rotation matrix \( R_p \), and translation matrix \( d \):
\begin{equation}
    \begin{bmatrix}
        l_{x,i} \\
        l_{y,i} \\
        l_{z,i}
    \end{bmatrix}
    =
    s \cdot R_p \cdot
    \begin{bmatrix}
        l_x \\
        l_y \\
        l_z
    \end{bmatrix}
    + d
\end{equation}

For orientation, quaternions are converted into a rotation matrix \( R_i \), adjusted by a predefined rotation matrix \( R_o \), and converted back to quaternions.

The error due to \gls{mtp} latency is quantified as a weighted \gls{rmse} between operator motion \( \tilde{p}_i \) and the rendered virtual arm pose \( \tilde{p}_v \):
\begin{equation}
    e_v = \omega_1 \cdot \text{RMSE}_p(\tilde{p}_i, \tilde{p}_v) + \omega_2 \cdot \text{RMSE}_o(\tilde{p}_i, \tilde{p}_v)
\end{equation}
Similarly, the real-world execution error is computed as:
\begin{equation}
    e_r = \omega_3 \cdot \text{RMSE}_p(\tilde{p}_i, \tilde{p}_r) + \omega_4 \cdot \text{RMSE}_o(\tilde{p}_i, \tilde{p}_r)
\end{equation} 

These metrics reflect the system’s ability to maintain high-fidelity trajectory following despite the presence of communication delays, mapping inaccuracies, and interface limitations, thereby serving as task-specific \glspl{kpi} for real-time control quality.

\subsubsection{Open Tasks: Real-time 3D Scene Representations for Remote Inspection} 
In nuclear fusion decommissioning, remote inspection is essential due to radiation, contamination, and structural degradation, making manual inspection infeasible. Robotic systems are used to gather visual and geometric data from hazardous areas. Accurate 3D scene representations are crucial for damage assessment and planning tasks like remote cutting and welding. However, challenges such as poor lighting, reflective surfaces, and limited camera viewpoints complicate the generation of high-fidelity models. To address this, an open-ended task is designed where the operator guides a robotic arm to capture images from multiple perspectives within a fixed time. These images are processed using a NeRF-based algorithm to create a 3D model. The task evaluates the system's performance in latency-sensitive inspection scenarios, with model quality assessed using visual metrics like \gls{psnr}, \gls{ssim}, and \gls{lpips}. The goal is to assess how well the system maintains inspection fidelity under real-time constraints in high-risk environments. Specifically, let $\{ {\bf{I}}_v(t_i) \}$ be the set of images captured at viewpoints, where $t_i \in [0, T_{task}]$ represents the time at which each image is captured. The task is to generate a 3D model $\mathcal{M}$ from these images:
\begin{equation}
\mathcal{M} = \mathcal{F}_{\text{NeRF}}( \{ {\bf{I}}_v(t_i) \} )
\end{equation}
Thus the quality of the model is evaluated using the \glspl{kpi}:
\begin{equation}
\text{KPI}(\mathcal{M}) = \{ \text{PSNR}(\mathcal{M}), \text{SSIM}(\mathcal{M}), \text{LPIPS}(\mathcal{M}) \}
\end{equation}
These \glspl{kpi} collectively reflect the system's ability to deliver high-fidelity 3D scene representations under time and communication constraints, which is critical for effective and safe decision-making in industrial real-time Human-robot
interaction.

\section{Human-in-the-loop Model-Agnostic Meta-Learning}\label{sec:hitl-mr}

In this section, we introduce the \gls{hitl}-\gls{maml} algorithm, where a two-stage training process is included. In the first stage, we train the~\gls{maml} based on the collected input data from the operators, functioning as a preheater that converges on multiple pre-defined tasks~\cite{finn2017model}. In the second stage, we use a human-in-the-loop online training strategy, where the operator interacts with industrial Metaverse in real-time, controlling the input device to generate control commands, and receive visual feedback to complete pre-defined tasks. In this stage, the model converges further based on the user's personalised characters and behaviours.

\subsubsection{State}
The state in the $t$-th time slot is a combined vector of the latest min-max normalized input pose ${\bf{p}}(t)$ and the \gls{e2e} latencies for real-world robotic control and the visual feedback, denoted by ${\bf{s}}_{t} = [{\bf{s}}_{t}^{[1]},{\bf{s}}_{t}^{[2]}] = [{\bf{p}}(t), T_r(t), T_v(t)],\  {\bf{s}}^{[1]}_t = {\bf{p}}(t) \in [-1,1]^{1 \times 7}, \ {{\bf{s}}^{[2]}_t} = [ T_r(t), T_v(t)]  \in \mathbb{N}^{1 \times 2}$.
\subsubsection{Action}
In the $t$-th time slot, the action to be taken are the lengths of prediction horizons for real-world robotic arm control $H_r(t)$ and for visual feedback $H_v(t)$, denoted by ${{\bf{a}}_t} = [{{\bf{a}}^{[1]}_t},{{\bf{a}}^{[2]}_t}] = [H_r(t), H_v(t)]$.

\subsubsection{Instantaneous Reward} 
In the $t$-th time slot, given state ${\bf{s}}_t$ and action ${\bf{a}}_t$, the instantaneous reward is a weighted sum of \glspl{rmse}, denoted by
\begin{align}\label{eq: instanous reward}
    r({\bf{s}}_t, {\bf{a}}_t) =  {\bf{e}}_v(t) +  {\bf{e}}_r(t),
\end{align}
where ${\bf{e}}_r(t)$ is the \gls{rmse} for the real-world robotic control and ${\bf{e}}_v(t)$ is the \gls{rmse} for the visual feedback.

\subsubsection{Policy}

The policy $\pi_{\theta}(\mathbf{a}_t \mid \mathbf{s}_t)$ maps states to action probabilities with parameters $\theta$. It consists of multiple fully connected layers: two initial layers extract compact features from input $[\mathbf{p}(t), T_r(t), T_v(t)]$, followed by layers optimizing prediction horizons for real-world and visual controls. A final linear layer with Softmax generates distributions $\rho_t^{[1]} \in \mathbb{R}^{H_r}$, $\rho_t^{[2]} \in \mathbb{R}^{H_v}$. Actions $\mathbf{a}_t^{[i]}$ are sampled as:
\begin{align}
\rho_t^{[i]} =
\begin{pmatrix}
   \Pr\{a_t^{[i]}=H\} \\ 
   \vdots \\ 
   \Pr\{a_t^{[i]}=1\} \\
   \Pr\{a_t^{[i]}=0\}
\end{pmatrix}, \quad i=\{1, 2\},
\end{align}
where $H$ is the maximum prediction length.

\subsection{Preliminary of MAML}\label{intro-maml}

\gls{maml} is a widely used algorithm in meta-learning, where the goal is to quickly adapt a model to new tasks with limited data~\cite{finn2017model}. The idea is that the convergence of the model on multiple tasks can be transformed into an aggregation that converges on each task. In our algorithm design, we use an extended version of its reinforcement learning to make it generalizable across multiple predefined real-time interactive tasks, as well as allowing for rapid adapting to the second stage online learning. During the process, the model is trained to be able to adapt to a large or infinite number of tasks. Here, we denote that there are $N \in \mathbb{N}^{+}$ tasks, ${\bf{T}} = {T_1, T_2, \dots, T_{N}}$, and each task has finite length $L$. For example, The $n$-th task $T_n$ can be defined as
 \begin{align}
   T_n = \{ \mathcal{L}_{T_n}(\pi_{\theta}), \{ ({\bf{a}}_{t_1}, {\bf{s}}_{t_1}), ..., ({\bf{a}}_{t_L}, {\bf{s}}_{t_L})\}, \rho_t, L \},
\end{align}
where $\{({\bf{a}}_{t_1}, {\bf{s}}_{t_1}), ..., ({\bf{a}}_{t_L}, {\bf{s}}_{t_L})\}$ represents a sequence of $L$ state-action pairs generated by following policy $\pi_{\theta}$, $\rho_t$ is the transition distribution of actions. 

Here, we adapt the loss function in \gls{maml} by incorporating the \gls{ppo} loss function~\cite{ppo},
\begin{align}\label{vf new}
\mathcal{L}_{T_n}(&\pi_{\theta}) = \min \left(\frac{\pi_{{\theta}}({\bf{a}}_t \mid {\bf{s}}_t)}{\pi_{\theta^k}({\bf{a}}_t \mid {\bf{s}}_t)}{A^{\pi_\theta}({\bf{s}}_t,{\bf{a}}_t)},\right. \\ \notag
&\left.\text{clip}\left(\frac{{\pi_{{\theta}}}({\bf{a}}_t \mid {\bf{s}}_t)}{\pi_{\theta^k}({\bf{a}}_t \mid {\bf{s}}_t)}, 1-\epsilon, 1+\epsilon\right)A^{\pi_{\theta_t}}({\bf{s}}_t,{\bf{a}}_t)\right),
\end{align}

where $\theta^k$ represents the parameters of policy network in previous $k$ steps, $A({\bf{s}}_t,{\bf{a}}_t)$ is the advantage function which estimates the advantage of taking action ${\bf{a}}_t$ in state ${\bf{s}}_t$, over other possible actions in the same state~\cite{ppo}, 
\begin{align}\label{eq: advantage}
A^{\pi_{\theta}}({\bf{s}}_t,{\bf{a}}_t) = Q^{\pi_{\theta}}({\bf{s}}_t,{\bf{a}}_t) - V^{\pi_{\theta}}({\bf{s}}_t),
\end{align}
 where $Q^{\pi_\theta}({\bf{s}}_t,{\bf{a}}_t)$ is the state-action value function and $V^{\pi_{\theta}}({\bf{s}}_t)$,is the state-value function. Then, based on~(\ref{vf new}), we compute the adapted parameters with one step gradient descent by each time sampling $K$ trajectories under policy $\pi_{\theta_n}$ for task $T_n$,
\begin{align}\label{eq: one gradient update}
\theta_{n} = {\theta} - \alpha \cdot \nabla_{{\theta}}\mathcal{L}_{T_n}(\pi_{\theta}),
\end{align}
where $\alpha$ represents the step length and $\nabla_{\theta}$ is the gradient with respect to the parameter ${\theta}$. 
After performing (\ref{eq: one gradient update}) for each task ${\bf{T}} = {T_1, T_2, \dots, T_{N}}$, we then implement the meta-optimization via \gls{sgd}, where the $\theta$ are further updated as follows,

 \begin{align}\label{eq: SGD}
{\theta}' \leftarrow {\theta} - \beta \cdot \nabla_{{\theta}}\sum_{n=1}^{N}\mathcal{L}_{T_n}(\pi_{\theta_n}),
\end{align}
where $\beta$ represents the meta-step length. 
\subsection{Problem Formulation}
We optimize the two prediction horizons for functions 1) real-world robotic arm control, and 2) visual feedback, to minimize the weighted sum of 1) the \gls{rmse} between the operator's input pose and the end-effector of the robotic arm $e_r(t)$, and 2) the \gls{rmse} between the operator's input pose and the pose of the \gls{dt} of the robotic arm displayed on the screen $e_v(t)$ in $N \in \mathbb{N}^{+}$ tasks, ${\bf{T}} = {T_1, T_2, \dots, T_{N}}$. For this purpose, we formulate it as a meta-reinforcement learning problem to find the optimal policy $\pi_\theta^*$,  
\begin{align}   
    \mathop{\arg\min}\limits_{\theta,H_v(t),H_r(t)} & \mathbb{E}_{{T_n}\sim{\bf{T}}}[\mathcal{L}_{T_n}(\theta - \alpha \cdot \nabla_{\theta}\mathcal{L}_{T_n}(\theta))] \label{eq: pf}\\
    \textbf{s.t.} \ \ \ \ \ \ \ \ \ \ & \notag \\
    \mathcal{L}_{T_n}(\theta)= \mathbb{E}&_{{\pi_{\theta},{T_n}}}[\sum_{t=0}^{L-1}\gamma^{t} r_n({\bf{a}}_{t_1}, {\bf{s}}_{t_1})], \forall{T_n} \sim  {\bf{T}} \tag{\ref{eq: pf}{a}}\\
&\ \ \ \ \ 0<H_v(t) < H_v \tag{\ref{eq: pf}{b}} \\
 &\ \ \ \ \  0<H_r(t) < H_r \tag{\ref{eq: pf}{c}}
\end{align}

where $\gamma$ represents the discount factor, $H_v$ denotes the maximum prediction horizon for the rendering loop and $H_r$ is the maximum prediction horizon for the control loop. 

\subsection{Two-Stage Training Process} 
\begin{figure}
            \centering
            \includegraphics[scale=0.365]{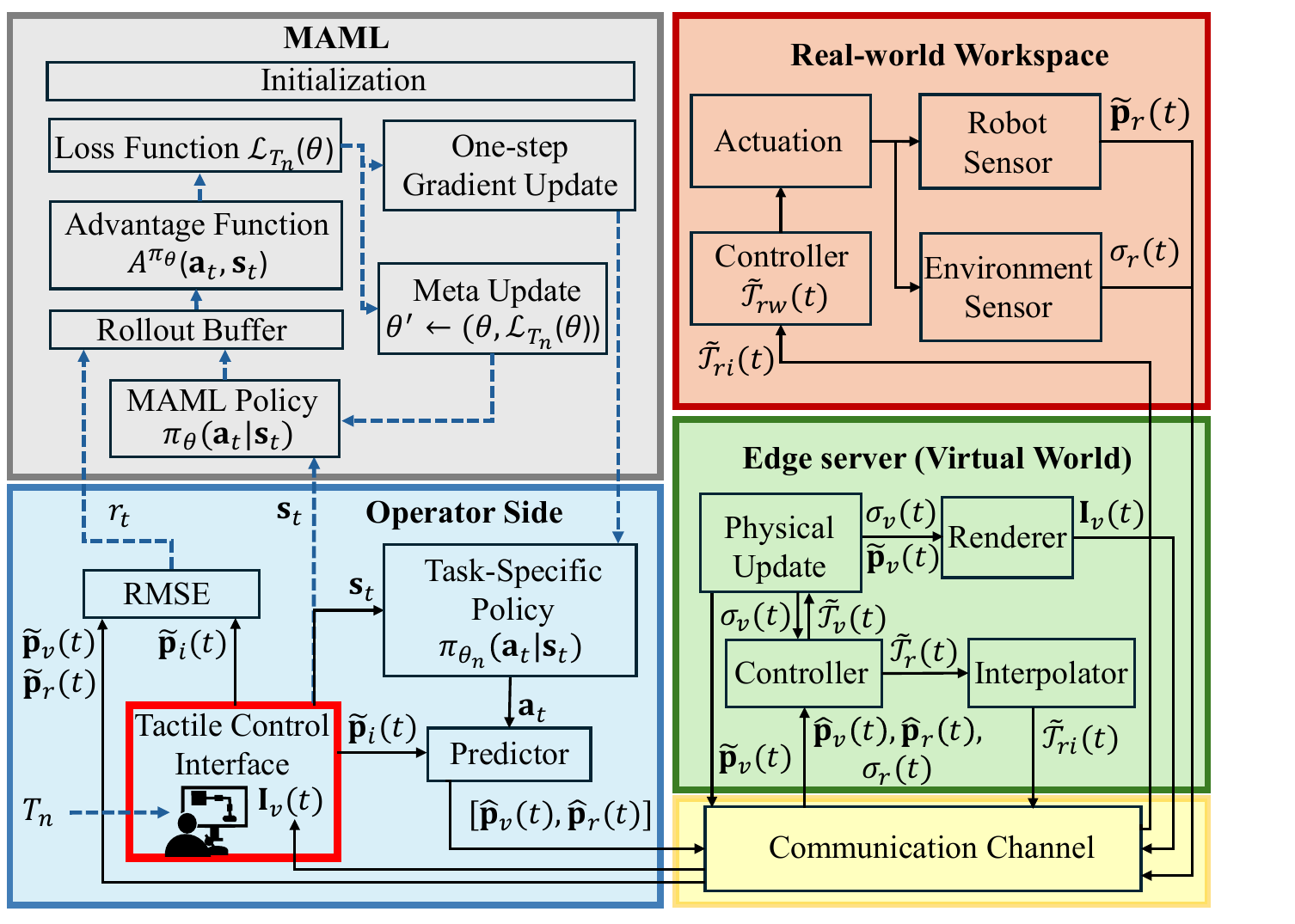}
          \caption{Illustration of proposed \gls{hitl} online training architecture.}
          \label{eq: Illustration of hitl transfer}
\end{figure}
Our proposed two-stage training algorithm is shown in Algorithm~\ref{algorithm}. In the first stage, we train the ~\gls{maml} based on the collected data which is detailed in Section~\ref{data collection}. The goal of this stage serves as a warm-up phase, which enables the agent to quickly acquire a policy for the next stage of training, i.e., real-time interaction using only a small amount of experience. In the second stage, \gls{hitl} online training is shown in Fig.~\ref{eq: Illustration of hitl transfer}. The operator interacts with a reinforcement learning environment - manipulating an input device in real time based on visual feedback to control a real-world robotic arm to perform a new interaction tasks. In this process, states and actions are generated online and reward is computed in real time. As the interaction continues, the model is trained until it converges.

\begin{algorithm}[t]\label{alg.l1}
\begin{algorithmic}[1]
\renewcommand{\algorithmicrequire}{\textbf{Input:}} 
\renewcommand{\algorithmicensure}{\textbf{Output:}} 

\caption{HITL-MAML} \label{algorithm}
\REQUIRE
initial all the parameters of neural network including randomly  initial $\theta_0$, initial state $\mathbf{s}_{0}$, step length $\alpha$, meta-step length $\beta$, Episode number for two stages $M_1,M_2$

\textbf{Stage 1:}
\WHILE{not done}
    \STATE Sample batch of tasks $T_n \in \bf{T}$
    \FOR{all $T_n$}
        \STATE Sample trajectories $\mathcal{D} = \{ ({\bf{a}}_{t_1}, {\bf{s}}_{t_1}), ..., ({\bf{a}}_{t_L}, {\bf{s}}_{t_L})\}$ using $\pi_{\theta}$ in $T_n$
        \STATE Evaluate $\nabla_{\theta}\mathcal{L}_{T_n}(\pi_\theta)$ using $\mathcal{D}$ and $\mathcal{L}_{T_n}(\pi_\theta)$ in~\eqref{vf new}
        \STATE Compute adapted parameters with gradient descent: $\theta_{n} = \theta - \alpha \cdot \nabla_{\theta}\mathcal{L}_{T_n}(\pi_\theta)$
        \STATE Sample trajectories $\mathcal{D}_n' = \{ ({\bf{a}}_{t_1}, {\bf{s}}_{t_1}), ..., ({\bf{a}}_{t_L}, {\bf{s}}_{t_L})\}$ using $\pi_{\theta+1}$ in $T_n$
    \ENDFOR
\STATE Update $\theta' \leftarrow \theta - \beta \cdot \sum_{n=1}^{N}\mathcal{L}_{T_n}(\pi_{\theta_n})$ using each $\mathcal{D}_n'$ and $\mathcal{L}_{T_n}(\pi_{\theta_n})$
\ENDWHILE

\textbf{Stage 2:}
\WHILE{not done}
        \STATE Operators interact with the new~\gls{drl} environment 
        \STATE Generate trajectories $\mathcal{D} = \{ ({\bf{a}}_{t_1}, {\bf{s}}_{t_1}), ..., ({\bf{a}}_{t_L}, {\bf{s}}_{t_L})\}$ based on $\pi_{\theta}$ and store in $T_n'$
        \STATE Repeat step 1-10 based on new tasks $T_n'$
\ENDWHILE
\ENSURE
Optimal policy ${\pi_{\theta}^{*}}$ with parameters $\theta^{*}$
\end{algorithmic}
\end{algorithm}

\section{Prototype Design}
\label{sec:prototype}

\subsection{System Setup}
\subsubsection{Input Device}
\begin{figure}
            \centering
            \includegraphics[scale=0.338]{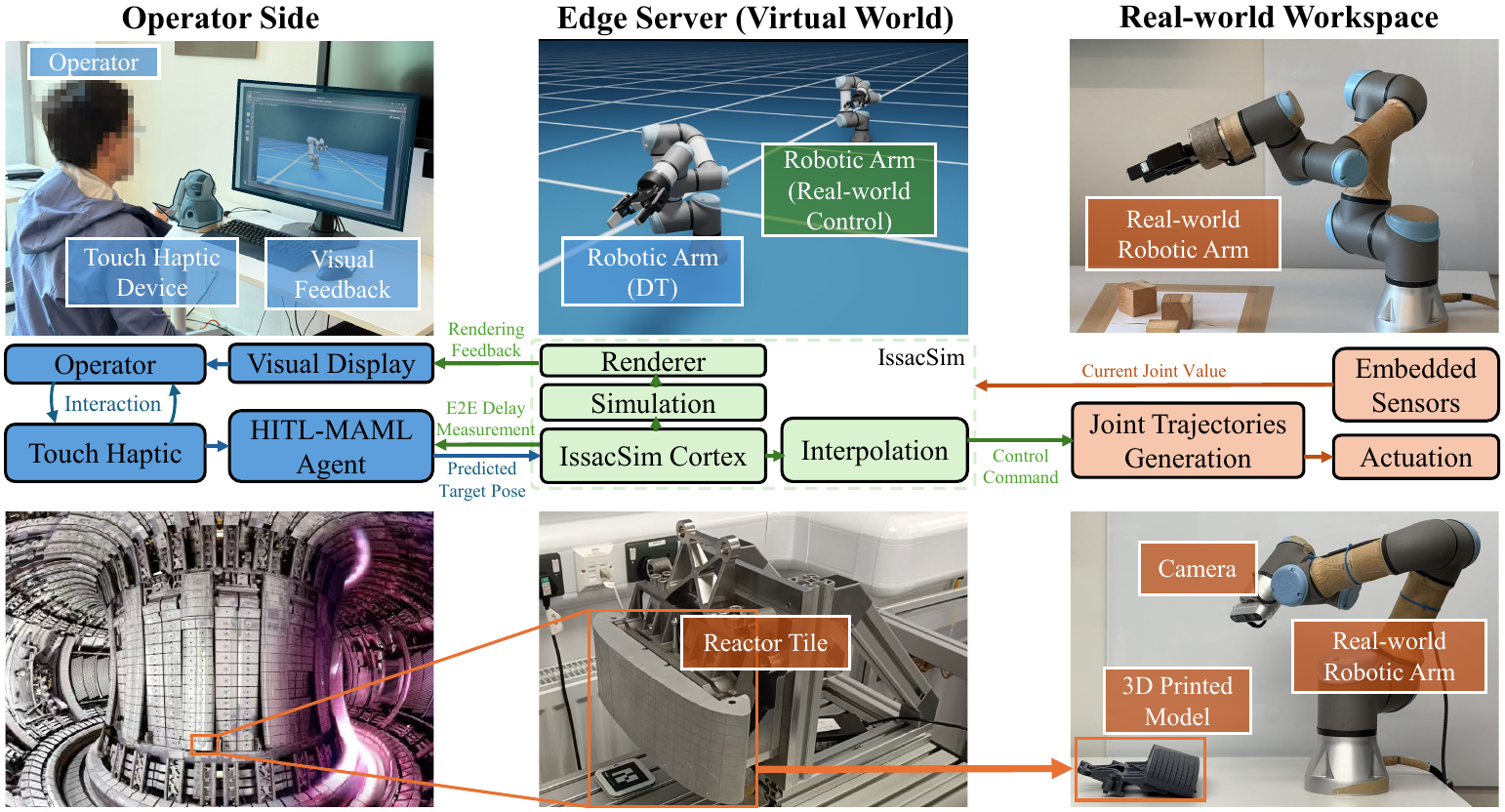}
           \caption{Illustration of our prototype design. 1) Upper left: A Touch Haptic device and the screen for visual display, 2) Upper central: A \gls{dt} of the robotic arm and the virtual environment deployed on the edge server, 3) Upper right: robotic arm in the real-world workspace, 4) Lower left:  nuclear fusion reactor, 5) Lower central: Split reactor tile, 6) Lower right: 3D scene scenario. 
           }
           \label{fig: prototype}
  \end{figure}

As shown in Fig.~\ref{fig: prototype}, the Touch Haptic Device is employed as the primary input mechanism, enabling the operator to seamlessly interact with the system~\cite{haptic}. This device provides six \gls{dof} for position and orientation sensing, alongside three degrees of force feedback. These capabilities offer an intuitive and immersive control interface, allowing for a tactile interaction experience. With a positional accuracy of up to 0.055~mm and a force feedback resolution as fine as 0.03~N, the device ensures precise input tracking, making it well-suited for tasks that require meticulous control. This high level of accuracy facilitates the fine manipulation of virtual objects, contributing to a smooth and responsive interaction between the operator and the Metaverse. In the context of our framework, the Touch Haptic Device enhances the operator’s ability to perform high-precision tasks with minimal error. Its integration into the system is crucial for scenarios demanding delicate and accurate manipulation, thus improving the overall functionality and efficiency of the framework.

\subsubsection{Visual Feedback}
We utilize a Dell G3223Q 4K monitor displaying the virtual work environment hosted on the edge server~\cite{dell_monitor}. This monitor provides high-resolution visual feedback of the \gls{dt} of the robotic arm and the Metaverse, with a refresh rate of 165 Hz, ensuring smooth and uninterrupted viewing. This allows the operator to assess the extent of the task based on visual feedback and make timely control via the touch device. Thus, the combination of control and visual feedback enhances the overall interaction, ensuring optimal performance in the proposed industrial Metaverse.
 
\subsubsection{The Metaverse}
The Metaverse is built on the NVIDIA Omniverse platform, utilizing Isaac Sim to simulate and control robotic systems~\cite{sim}, which is running on a high-performance edge server equipped with 32 GB RAM, an NVIDIA RTX 3070 GPU, and an Intel Core i7-11700 16-core CPU. We employ Isaac Sim to create a highly accurate \gls{dt} of the real-world robotic arm, modeling its dimensions and functionalities in detail. In addition to the physical simulation, our framework integrates Isaac Cortex, which enhances task awareness and allows for more advanced decision-making capabilities in robotic systems~\cite{isaac_cortex}. Instead of merely simulating physical interactions, our framework uses this decision-making layer to optimize how robots approach and complete tasks, leading to more efficient operations. We also decouple the physics updates from visual rendering in Isaac Sim, allowing us to maintain high-frequency physical updates while ensuring smooth visual feedback on the operator's monitor. This setup ensures that the operator receives continuous real-time feedback while controlling the virtual environment, ensuring accurate task execution. The overall framework, with its integration of Isaac Sim and Isaac Cortex, creates a tightly coupled system between the operator side, the Metaverse, and the real-world robotic arm, enabling precise and efficient task management. 

\subsubsection{Real-World Robotic Arm}
In our framework, we employ the Universal Robots UR3e industrial robotic arm as the core component for executing physical tasks~\cite{ur3e}. The UR3e offers six \gls{dof} and achieves a repeatability of ±0.03 mm, making it well-suited for tasks that require high precision. With a payload capacity of 3 kg and a reach of 500 mm, the UR3e is optimized for small-scale, detailed operations while maintaining flexibility in diverse working environments. In the real-world workspace, the UR3e interacts with objects that are digitally mirrored in the Metaverse, ensuring seamless integration between the cyber and physical domains. After receiving the control command from the edge server, the UR3e implements the control commands by utilizing the MoveIt Kinematics Plugin~\cite{ros_moveit}, which enables smooth and accurate joint motion execution. In addition, during the entire working process, the built-in sensors of the robotic arm monitor the joint angles, angular velocities, and applied forces to maintain consistent and precise control. This sensing system is critical for maintaining the accuracy required in our task-oriented framework, ensuring that every command is executed with the necessary precision and accurate task-oriented \glspl{kpi} calculations.

\subsubsection{Networks}
Our framework implements a publish-and-subscribe mechanism in the \gls{ros} environment to facilitate communication between the operator, the edge server, and the real-world workspace~\cite{ros}. This mechanism allows for flexible and scalable data exchange, enabling the modular development of complex cyber-physical applications. Communication between processes relies on two protocols: \gls{rpc} and \gls{tcp}~\cite{ros_tcp}. When transmitting data, the local process first involves a publisher (talker) registering with the \gls{ros} Master on a designated port, including the topic name in its registration. A subscriber (listener) then registers with \gls{ros} Master and requests information based on its subscription. The \gls{ros} Master, upon finding a matching topic publisher, provides the listener with the talker’s \gls{rpc} address. The listener sends an \gls{rpc} request to the talker, initiating the connection by specifying the topic name and message type. Once the talker confirms this request, it provides its \gls{tcp} address, and the listener establishes a network connection. The talker can then begin publishing data to the listener over \gls{tcp}. To simulate and control communication delays within this network, we used the network cables to connect the three via Ethernet during training, making the communication delay of the benchmarks nearly zero. Then, we utilize the \gls{tc} method in Linux~\cite{tc}. This utility allows us to manipulate network delay by introducing artificial delays, enabling controlled experiments on how varying levels of network delay affect the system's performance. By systematically varying communication delays with \gls{tc}, we can observe their impact on the control and response of the robotic system. These controlled network conditions allow us to generate reproducible results for training and testing.

\begin{table*}[t]\label{parameters}
\renewcommand{\arraystretch}{1}
\centering
\caption{System Parameters for Performance Evaluation}
\begin{threeparttable} 
\resizebox{!}{!}{
\begin{tabular}{|l|l|l|l|}
\hline 
\multicolumn{2}{|c|}{\textbf{Prototype Setup}}  & \multicolumn{2}{c|}{\textbf{Learning Setup}} \\ 
\hline
\bf{Parameters} & \bf{Values} &\bf{Parameters} & \bf{Values}\\
\hline
Input device sampling frequency& \SI{120}{Hz} 
& Step length of one-step gradient update $\alpha$ & $\num{1e-3}$\\
\hline
Input length of prediction function $W_p$ & \SI{4000} {\milli\second}
& Meta-step length $\beta$ & $\num{1e-5}$\\
\hline
Prediction horizon of prediction function $H$ & \SI{1000}{\milli\second}  
& Batch size& {256} \\
\hline
Simulation frequency ${1}/T_{si}$ & \SI{240}{Hz}
&Discount factor $\gamma$ & $\num{0.99}$\\
\hline
Rendering step length $T_{im}$ & \SI{16}{\milli\second} 
& General advantage estimation parameter $\lambda$  & $\num{0.99}$\\
\hline
Rendering resolution & $1920\text{px} \times 1080\text{px}$ 
& Clip range  & 0.2\\
\hline
Real-world robotic arm sampling frequency & \SI{1000}{Hz} &
Weighting coefficient $\omega_1,\omega_2,\omega_3,\omega_4$ & {-1} \\
\hline 
\end{tabular}
}
\end{threeparttable}
\label{tab:sys_param}
\end{table*}

\begin{figure}
            \centering
           \includegraphics[scale=0.16]{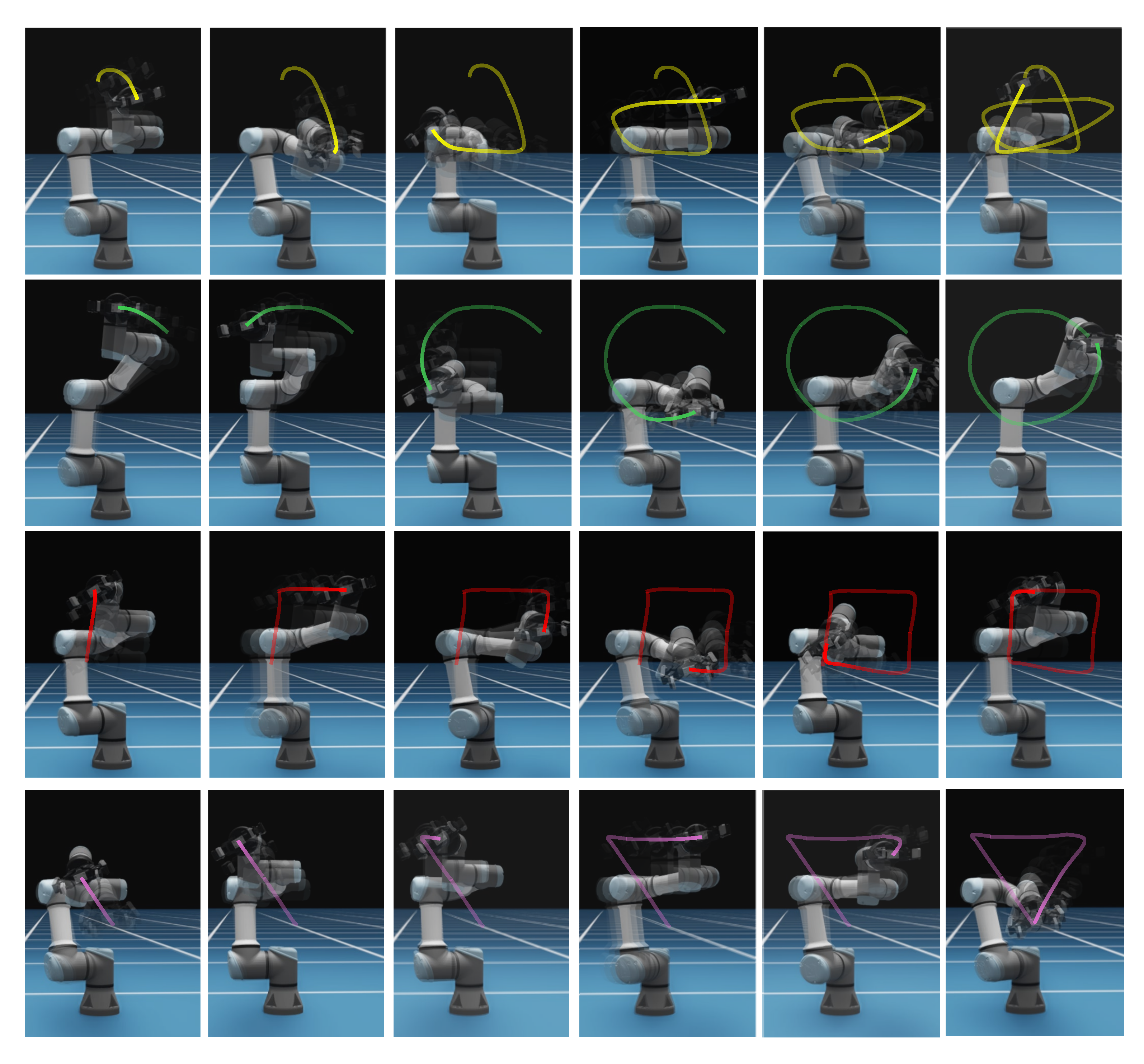}
           \caption{Illustration of four real-time interaction drawing shapes, i.e., pentagram, circle, square, and triangle.}
           \label{fig: traj}
\end{figure}

\subsection{Data Collection}\label{data collection}
As shown in fig.~\ref{fig: traj}, we collected real-time interaction data from four trajectory-following tasks, where an operator used a touch haptic device to control a robotic arm through shapes including circle, triangle, pentagram, and square. These trajectories are designed to simulate typical motion patterns in nuclear decommissioning tasks, such as remote cutting, surface following, and precise manipulation of in-vessel components~\cite{kayan2022cybersecurity}. The operator received real-time 3D visual feedback from a digital twin, enabling on-the-fly adjustments to trajectory parameters (e.g., shape, size, and speed) to adapt to dynamic task requirements and workspace constraints. Both the target pose from the haptic device and the actual pose of the robot end-effector were continuously recorded in \gls{csv} format. Data collection consisted of two phases: (1) 150 repetitions of each trajectory shape for pretraining, simulating repetitive precision operations, and (2) online learning during novel, task-specific executions until model convergence. This dataset forms the foundation for training and validating our predictive control framework under realistic remote operation conditions relevant to nuclear inspection and maintenance.

\subsection{Validation in Remote Inspection Tasks for Nuclear Decommissioning}\label{Industrial task scenarios}
In addition to implementing \gls{hitl}-\gls{maml}  online training on predefined trajectories, we validate the effectiveness of \gls{hitl}-\gls{maml} in a real-world industrial application focused on high-precision 3D visualization and inspection of reactor tiles in nuclear decommissioning. For evaluation, we use full-scale 3D-printed replicas of reactor tiles. Expert operators remotely control the robotic arm along predefined paths, pausing at strategic points using the haptic touch interface to capture images. The 3D scene representation server generates 3D models from these images and camera poses using a NeRF-based algorithm, enabling realistic remote inspection with novel view synthesis~\cite{pachecoimportance, mueller2022instant}. However, due to the communication delays inherent in remote control, operators experience noticeable lag, which can hinder their ability to control the robotic arm precisely within a limited timeframe. This time lag can affect the accuracy of the arm’s movements and the overall inspection process~\cite{rakita2020effects}. To address these challenges, we apply \gls{hitl}-\gls{maml}, assessing its effectiveness in improving control precision and 3D inspection accuracy under latency constraints. Experimental details will be provided in the evaluation section.

\section{Results}\label{sec:results}
\subsection{Training Settings}

Algorithm training was conducted on a server with an Intel Core i9-13900F CPU, 64 GB RAM, and an NVIDIA RTX 4090 GPU. The default parameters of the prototype and learning algorithm are listed in Table~\ref{tab:sys_param}, unless otherwise specified. The parameters of \gls{hitl}-\gls{maml} are updated following the procedure outlined in Algorithm~\ref{algorithm}. In the first stage, we use the collected motion data mentioned in Section~\ref{data collection}. The framework sequentially executes each drawing shape over $4 \times 10^4$~ms. In the second stage, the operator interacts with the \gls{drl} environment: 1) \gls{hitl} training for incorporating human operation;2) 3D scene representations for remote inspection. The training time of each drawing shape is set to $2 \times 10^4$~ms. To simulate network delays $T_d$, a normal distribution with a mean of $\mu_c = 50$~ms and a standard deviation of $\sigma_c = 10$~ms is set between the operator side and the edge server.  

\subsection{Performance Evaluation of the Prediction Model}
\begin{figure}
            \centering
           \includegraphics[scale=0.4]{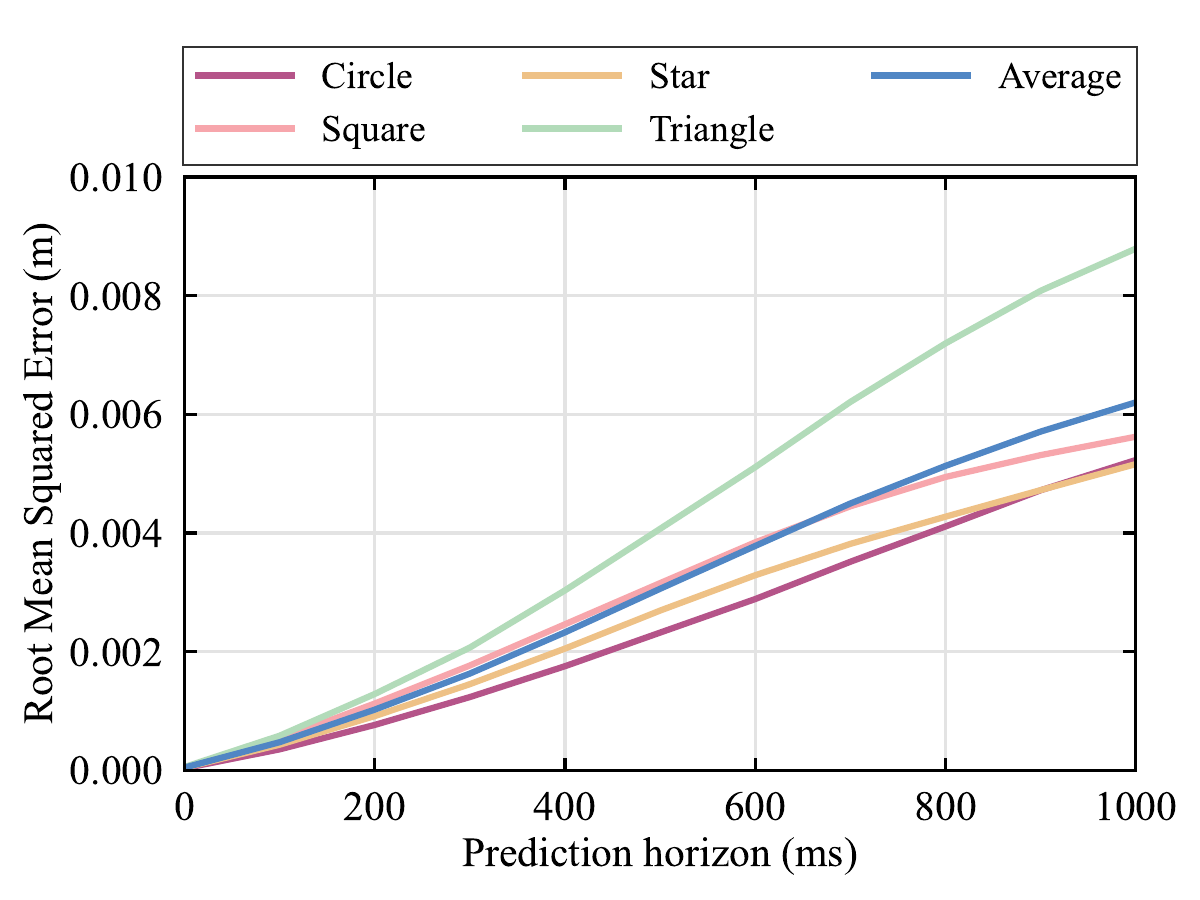}
           \caption{Evaluation of prediction model.}
           \label{fig: pre}
\end{figure}
As shown in Fig.~\ref{fig: pre}, the performance of the predictor \gls{arma} is evaluated by the \gls{rmse} with different prediction horizons. The results show that the predictor achieves similar performance in four predefined trajectories, with less than 0.01m \gls{rmse} when the prediction horizon is less than 1000~ms.
It is intuitively that as the prediction horizon increases, the prediction error also rises. Thus, there exists an optimal point where there is neither an increase in error due to prediction by compensating for too much delay, nor an insufficient prediction where there is still an error. Establishing this level of prediction accuracy is crucial, as it enables robust performance when dynamically adjusting prediction horizons during interactions with the \gls{drl} environment.

\subsection{Performance Evaluation of Trajectory Tracking Tasks-Drawing Shapes}

\subsubsection{Performance on the Collected Dataset}

\begin{figure}
            \centering
           \includegraphics[scale=0.4]{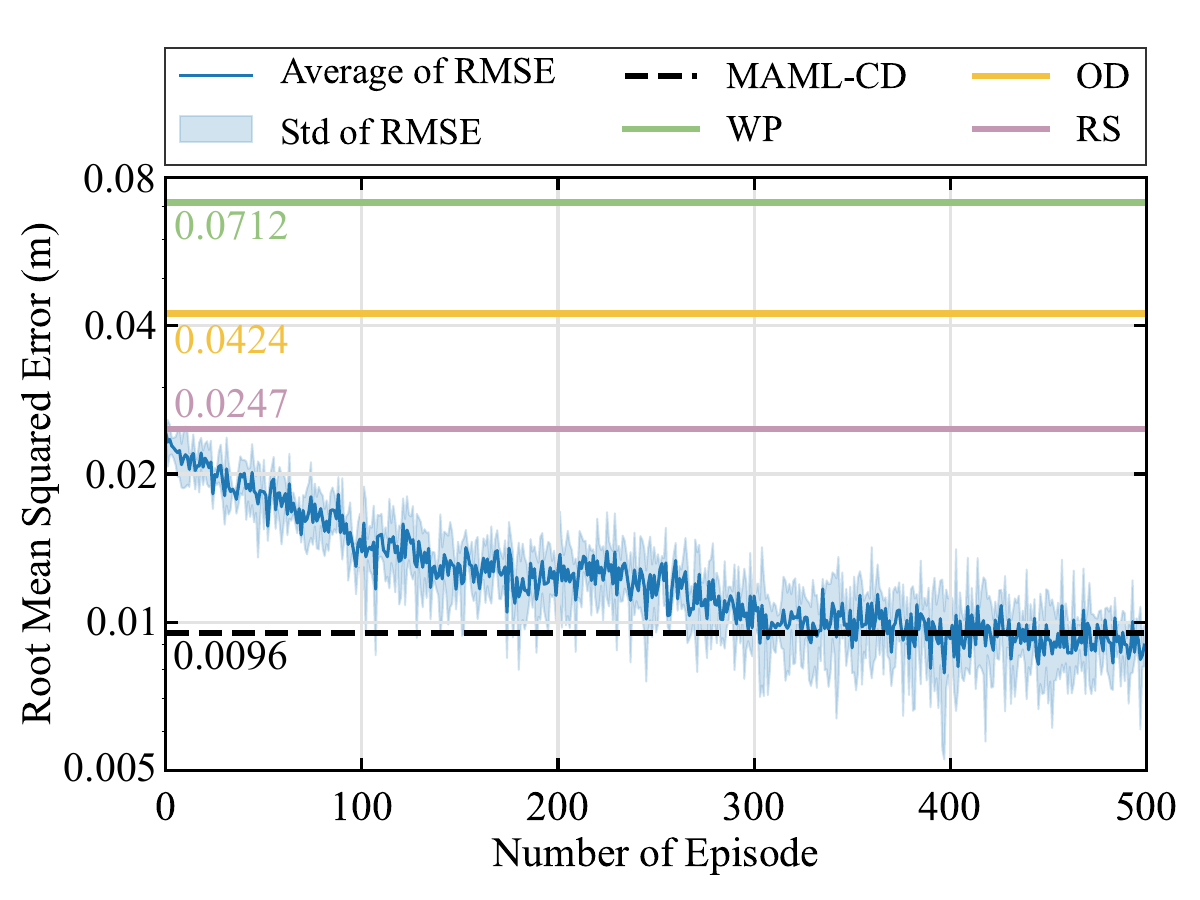}
           \caption{Evaluation of \gls{maml}-CD compared to three baselines.}
           \label{fig: maml}
\end{figure}

As shown in Fig.~\ref{fig: maml}, we train the \gls{maml} algorithm over 500 episodes for three times to evaluate the proposed \gls{maml} training process on the collected dataset.  To better illustrate our algorithm in the figure, we denote the \gls{maml} training process on the collected dataset as "{\gls{maml}-CD}". The \gls{rmse} gradually decreases during the first 300 episodes and achieves convergence at an \gls{rmse} of $0.0096$~m after 350 episodes of training. ``Std" represents the standard deviation. To verify the effectiveness of the proposed algorithm, We compare the performance of the trained \gls{maml} with three baselines,
\begin{itemize}
    \item {WP}: Without any delay compensation strategy, the pose of the input device is directly transmitted.
    \item {RS}: We predict the pose for the two functions, where the prediction horizons $H_r(t)$ and $H_v(t)$, are randomly sampled within the maximum value of $H$. 
    \item {OD}: The prediction horizons $H_r(t)$ and $H_v(t)$, are dynamically changed depends on the measured \gls{e2e} delays, $T_r(t)$ and $T_v(t)$. For example, if \gls{e2e} delay of real-world robotic control and visual feedback is 127ms and 133ms, respectively, we set $H_r(t)=127$ and $H_v(t)=133$.
\end{itemize}
The results show a significant superiority over the three baselines. The \gls{rmse} decrease by 249.39\%, 132.79\%, and 61.13\% compared to 'WP', 'OD', and 'RS', respectively.

\subsubsection{Performance of HITL Online Training}

\begin{figure}
            \centering
           \includegraphics[scale=0.4]{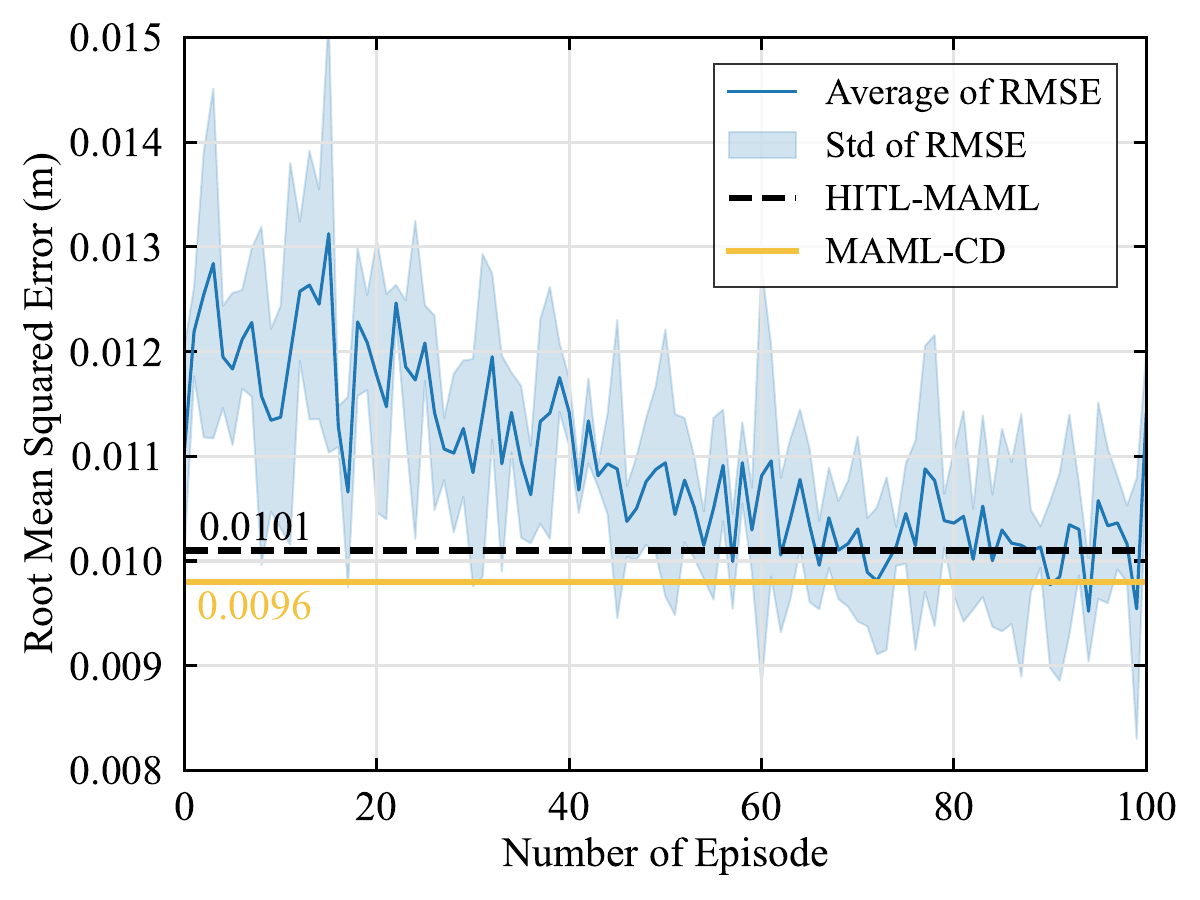}
\caption{Evaluation of Proposed \gls{hitl}-\gls{maml}.}
           \label{fig: hitl-maml}
\end{figure}

As shown in Fig.~\ref{fig: hitl-maml}, the performance of the \gls{hitl}-\gls{maml} training process is evaluated by incorporating human operators controlling the input device during training. we compare the results with \gls{maml}-CD and the other three baselines. The results show that \gls{hitl}-\gls{maml} rapidly converges after training for 80 episodes with an average \gls{rmse} of $0.0101$~m. which is close to the \gls{rmse} of \gls{maml}-CD. This indicates that our proposed algorithm can quickly adapt to the \gls{hitl} mode after a short period of online training, which verifies the effectiveness and efficiency of our algorithm. In addition, as shown in Tab.~\ref{teble_four_tasks}, we further examined the \gls{rmse} of the proposed algorithm in each task, and compared it to the three baselines mentioned above. The results show that the average \gls{rmse} of our proposed algorithm is less than that of the three baselines in each task, further illustrating the effectiveness of our algorithm. We get the lowest \gls{rmse} in task 'square', where the average \gls{rmse} of position and orientation are $0.0068$~m and $0.0089$, respectively.

\subsubsection{Evaluation in Different Communication Conditions}

\begin{figure}[t]
    \centering
    \vspace{2mm}
    \includegraphics[width = 0.48\textwidth]{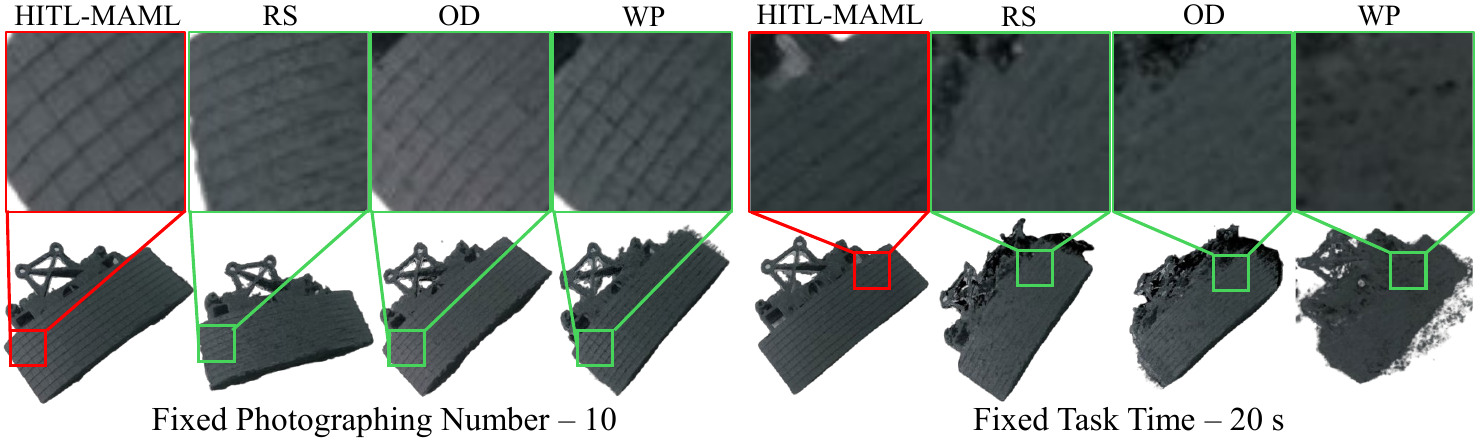}
    \caption{Evaluation of Performance for Industrial Task: Proposed algorithm vs. Three baseline policy.}
    \label{fig: Comapre_rendering}
\end{figure}

\begin{figure}
            \centering
           \includegraphics[scale=0.4]{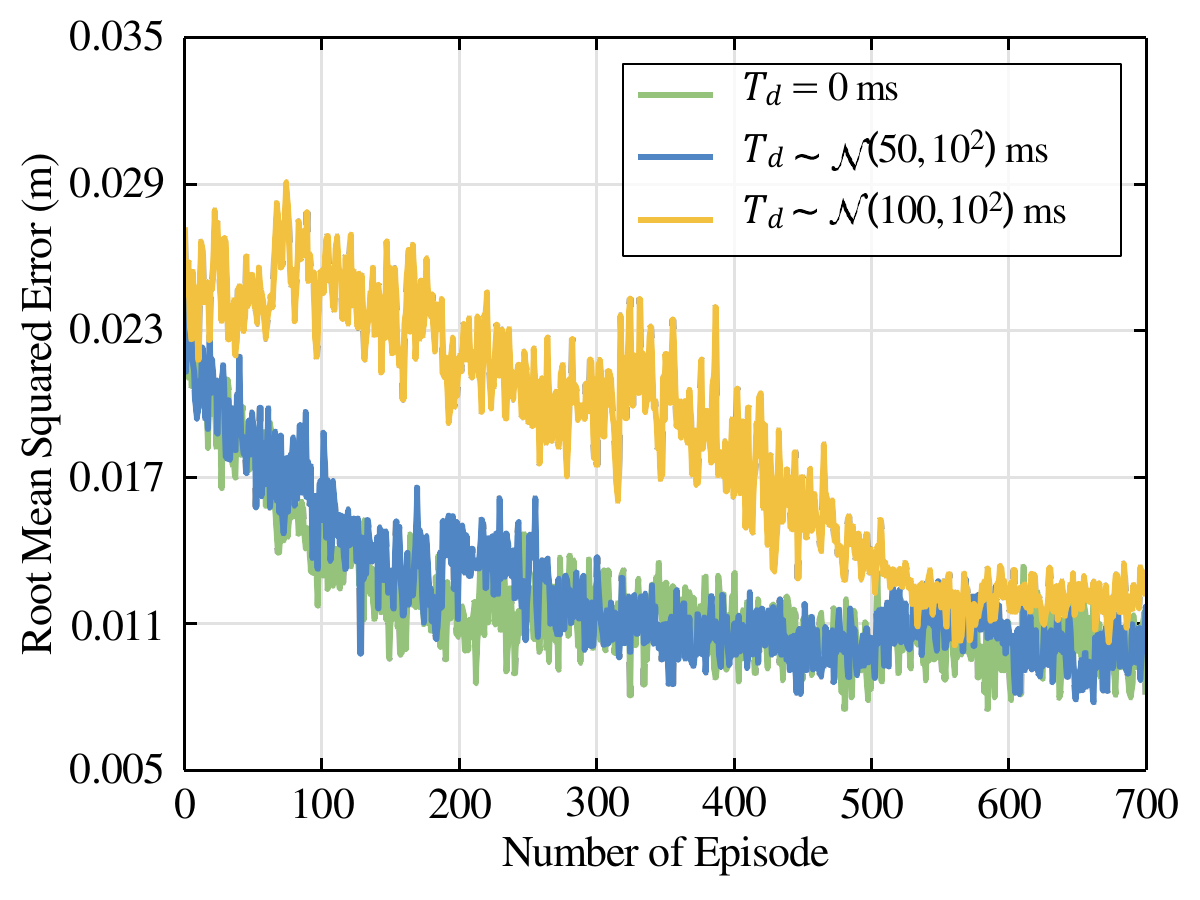}
           \caption{Evaluation of performance for different communications delays.}
           \label{fig: diff_delay}
\end{figure}

\begin{table}
\centering
\caption{Measured \gls{e2e} delays and the number of convergence episode}
\label{table_diff_delay}
\resizebox{0.99\linewidth}{!}{
\begin{tabular}{|c|c|c|c|}
\hline
  Communication delay (ms)& $T_d = 0$   & $T_d\sim\mathcal{N}(50,10^2)$ & $T_d\sim\mathcal{N}(100,10^2)$ \\ 
\hline
Average \gls{rmse} (m) & 0.0098  & 0.0094 & 0.01435  \\ 
\hline
Convergence episode number & 220  & 350 &  540 \\
\hline
Average \gls{e2e} delay (the Metaverse) (ms)& 29.1  & 127.4 &  236.2 \\
\hline
Average \gls{e2e} delay (real-world workspace) (ms)& 35.5  & 143.8 &  251.7 \\
\hline
\end{tabular}}
\end{table}

As shown in Fig.~\ref{fig: diff_delay}, We evaluate the performance of the proposed algorithm under different communication delay conditions. Specifically, In addition to the $0$~ms communication delay as a benchmark, we add different normally distributed delays, with mean $\mu_c = 50, 100$~ms and a standard deviation of $\sigma_c = 10$~ms is set between the operator side and the edge server. The results show that our proposed algorithm can successfully converge and reach an \gls{rmse} of approximately $0.011$ m. We also measured the \gls{e2e} as shown in Tab.~\ref{table_diff_delay}. The results also indicate that a higher \gls{e2e} delay needs to make the algorithm take more time to converge.

\begin{table}
\centering
\caption{Evaluation of Performance for different tasks}
\label{teble_four_tasks}
\resizebox{0.99\linewidth}{!}{
\begin{tabular}{|c|c|c|c|c|c|}
\hline
\gls{rmse} & Pose & \gls{hitl}-\gls{maml}   & "WP" & "OD" & "RS" \\ 
\hline
\multirow{4}{*}{\begin{tabular}[c]{@{}c@{}} Circle 
\end{tabular}}
& Position-vitual ($\text{m}$)  &  0.0091    &   0.0694  &  0.0398 & 0.0195\\ 
\cline{2-6}
& Orientation-vitual            &  0.0086    &  0.0475    & 0.0445  & 0.0223\\ 
\cline{2-6}
&Position-real ($\text{m}$)     &  0.0098   &  0.0812    & 0.0413 & 0.0264\\
\cline{2-6}
&Orientation-real            &  0.0104        & 0.0546   & 0.0476 & 0.0211\\
\hline
\multirow{4}{*}{\begin{tabular}[c]{@{}c@{}} Square
\end{tabular}}
& Position-vitual ($\text{m}$)  &  0.0067       & 0.0612  & 0.0365  & 0.0265\\ 
\cline{2-6}
&Orientation-vitual            &   0.0068    &  0.0461  & 0.0278 & 0.0189\\ 
\cline{2-6}
&Position-real ($\text{m}$)    &   0.0089  & 0.0689   & 0.0374  & 0.0210\\
\cline{2-6}
&Orientation-real            &  0.0089   &  0.0691   & 0.0377 & 0.0227\\
\hline
\multirow{4}{*}{\begin{tabular}[c]{@{}c@{}} Star
\end{tabular}}
& Position-vitual ($\text{m}$)  &   0.0099  & 0.0649   & 0.0514  & 0.0342\\ 
\cline{2-6}
&Orientation-vitual            &  0.0123   & 0.0832   & 0.0446 & 0.0258\\ 
\cline{2-6}
&Position-real ($\text{m}$)   &  0.0137   & 0.0792   & 0.0481 & 0.0298\\
\cline{2-6}
&Orientation-real            &  0.0121   &  0.0711  & 0.0475 & 0.0243\\
\hline
\multirow{4}{*}{\begin{tabular}[c]{@{}c@{}} Triangle 
\end{tabular}}
& Position-vitual ($\text{m}$)  &  0.0106   &  0.0715 &  0.0427 & 0.0279\\ 
\cline{2-6}
&Orientation-vitual            &  0.0114   &  0.0692  & 0.0392 & 0.0332\\
\cline{2-6}
&Position-real  ($\text{m}$)  &  0.0125   &  0.0811  & 0.0479 & 0.0295\\
\cline{2-6}
&Orientation-real            & 0.0118   &  0.0732  & 0.0401 & 0.0224\\
\hline
\end{tabular}}
\end{table}

\subsection{Performance Evaluation of pen Tasks: Real-time 3D Scene Representations for Remote Inspection}
\begin{table}
\centering
\caption{Comparative Analysis of Proposed algorithm Performance for Real-time 3D scene representation}
\label{table_length}
\resizebox{1\linewidth}{!}{
  \begin{tabular}{|c|c|c|c|c|c|c|c|c|}
    \hline  
     & \multicolumn{4}{c}{\textbf{Fixed Photographing number - 10}} & \multicolumn{4}{|c|}{\textbf{Fixed Task Time - 20 s}}\\
    \hline    
    \textbf{Approaches} & HITL-MAML & "WP" & "OD" & "RS" & HITL-MAML & "WP" & "OD" & "RS"\\
    \hline
    Average Task Time (s)& 20.4 & 34.2 & 29.6 & 28.1 & \multicolumn{4}{c|}{---------------------------------------------}\\
    \hline
     Average Photo Taken & \multicolumn{4}{c|}{---------------------------------------------}& 10 & 5 & 7 & 7 \\
    \hline
    Average RMSE (m) & 0.0116 & 0.0639 & 0.0395 & 0.0312& 0.0120 & 0.0758 & 0.0413 &0.0298\\
    \hline
    PSNR & 22.11 & 21.94 & 22.03 & 22.49 & 23.81 & 19.46 & 23.21 & 22.84\\
    \hline
    SSIM & 0.8729 & 0.8465 & 0.8647 & 0.8701 & 0.8854 & 0.8011 & 0.8753 & 0.8677 \\
    \hline
    LPIPS & 0.1298 & 0.1443 & 0.1340 & 0.1295 & 0.1171 & 0.2007 & 0.1233 & 0.1341\\
    \hline
  \end{tabular}
}
\end{table}
To evaluate the effectiveness of our proposed \gls{hitl}-\gls{maml} algorithm in the context of 3D scene representation for remote nuclear inspection, we designed two task-oriented experimental settings:
(1) Fixed Viewpoint Budget: the operator was allowed a fixed number of image captures, and the evaluation focused on task completion time and representation quality;
(2) Fixed Time Budget: the operator had a fixed time to explore the environment, and the number of captured viewpoints and representation quality were measured.
The quality of Real-time 3D scene representation was assessed both visually (Fig.~\ref{fig: Comapre_rendering}) and quantitatively using \gls{psnr}, \gls{ssim}, and \gls{lpips} metrics. A professional operator conducted the image-capturing task by remotely controlling the robotic arm using our \gls{hitl}-\gls{maml} framework and three baseline methods. As summarized in Table~\ref{table_length}, our method consistently achieved higher representation fidelity and reduced task time, particularly under time-constrained conditions. These results demonstrate the practical advantage of our approach for time-sensitive and communication-constrained inspection tasks.

\subsection{Demonstration of RMSE}
\begin{figure}
            \centering
           \includegraphics[scale=0.75]{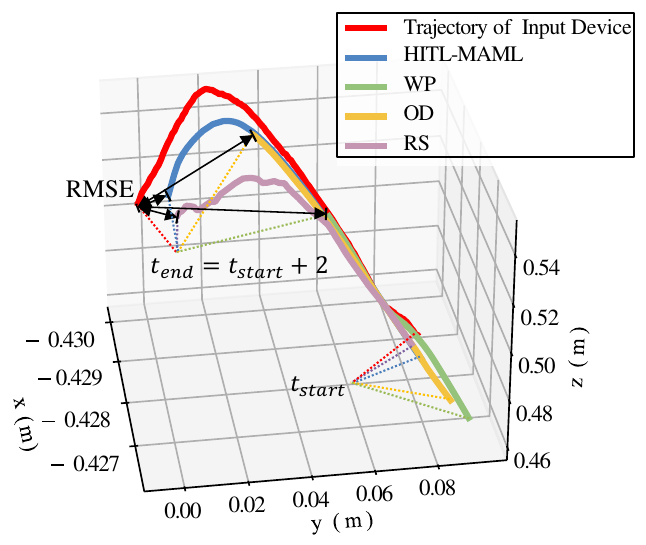}
\caption{Demonstration of the trajectories of the end effector.}
           \label{fig: ee}
\end{figure}

As shown in Fig.~\ref{fig: ee}, we demonstrate the~\gls{rmse} by showing different trajectories for two seconds. The input device's trajectories are transformed into the robotic arm's coordinate system. 
Due to the fact that the input device is the one that issued the command, it is logical that it is ahead of the other trajectories at the start point. It can be visualized that our proposed algorithm achieves the lowest \gls{rmse} at the endpoint, with the point that is closest to that of the input device. This provides a more intuitive illustration of the algorithm’s effectiveness.

\section{Conclusions}
\label{sec:conclusions}

In this study, we proposed a task-oriented cross-system framework leveraging \glspl{dt} to facilitate real-time human-device interaction in industrial Metaverse. By decoupling DTs into visual display and robotic control components, the system optimized responsiveness and adaptability under strict computational and network constraints. The integration of the \gls{hitl}-\gls{maml} algorithm enabled dynamic adjustment of prediction horizons, improving the generalizability of operator motion prediction. Experimental evaluations demonstrated the effectiveness of the proposed approach: in the Trajectory-Based Drawing Control task, the framework significantly reduced the weighted RMSE from 0.0712 m to 0.0101 m; in the real-time 3D scene representation task for nuclear decommissioning, it achieved a \gls{psnr} of 22.11, \gls{ssim} of 0.8729, and \gls{lpips} of 0.1298. These results validated the framework’s capability to ensure both spatial precision and visual fidelity in real-time, high-risk industrial scenarios.

\bibliography{IEEEabrv,main}
\bibliographystyle{IEEEtran}


\end{document}